\documentclass[10pt,twocolumn,letterpaper]{article}

\usepackage{iccv}
\usepackage{times}
\usepackage{epsfig}
\usepackage{graphicx}
\usepackage{amsmath}
\usepackage{amssymb}

\usepackage{booktabs}
\usepackage{multirow}
\usepackage{makecell}

\usepackage[printonlyused,withpage]{acronym}
\usepackage{amsmath}
\usepackage{etoc}
\usepackage{bm,bbm}

\usepackage{color}
\usepackage{xcolor}
\usepackage{colortbl}
\usepackage{pifont}

\usepackage{listings}
\usepackage{algorithm}
\usepackage{algorithmic}
\usepackage{rotating}
\usepackage{booktabs} 
\usepackage{xparse}
\usepackage{lipsum}

\usepackage{subcaption}

\usepackage[accsupp]{axessibility}

\makeatletter
\newcommand\figcaption{\def\@captype{figure}\caption}
\newcommand\tabcaption{\def\@captype{table}\caption}
\makeatother

\usepackage{etoolbox}
\makeatletter
\AfterEndEnvironment{algorithm}{\let\@algcomment\relax}
\AtEndEnvironment{algorithm}{\kern2pt\hrule\relax\vskip-3pt\@algcomment}
\let\@algcomment\relax
\newcommand\algcomment[1]{\def\@algcomment{\footnotesize#1}}
\renewcommand\fs@ruled{\def\@fs@cfont{\bfseries}\let\@fs@capt\floatc@ruled
	\def\@fs@pre{\hrule height.8pt depth0pt \kern2pt}%
	\def\@fs@post{}%
	\def\@fs@mid{\kern2pt\hrule\kern2pt}%
	\let\@fs@iftopcapt\iftrue}
\makeatother

\usepackage[pagebackref=true,breaklinks=true,letterpaper=true,colorlinks,bookmarks=false]{hyperref}

\usepackage[capitalize]{cleveref}
\crefname{section}{Sec.}{Secs.}
\Crefname{section}{Section}{Sections}
\Crefname{table}{Table}{Tables}
\crefname{table}{Tab.}{Tabs.}

\iccvfinalcopy 

\def\iccvPaperID{1432} 

\ificcvfinal\fi

\begin{document}

\title{Revisiting Foreground and Background Separation in Weakly-supervised Temporal Action Localization: A Clustering-based Approach}

\author{Qinying Liu\textsuperscript{\rm 1}
\quad Zilei Wang\thanks{Corresponding author} \textsuperscript{\rm 1}
\quad Shenghai Rong \textsuperscript{\rm 1}
\quad Junjie Li  \textsuperscript{\rm 1} 
\quad Yixin Zhang\textsuperscript{\rm 1,2} \\
\textsuperscript{\rm 1} University of Science and Technology of China\\
\textsuperscript{\rm 2} Institute of Artificial Intelligence, Hefei Comprehensive National Science Center \\
\small{\texttt{\{lydyc, rongsh, hnljj@\}@mail.ustc.edu.cn, \{zhyx12, zlwang\}@ustc.edu.cn}}
}

\maketitle
\ificcvfinal\thispagestyle{empty}\fi

\begin{abstract}
Weakly-supervised temporal action localization aims to localize action instances in videos with only video-level action labels. Existing methods mainly embrace a localization-by-classification pipeline that optimizes the snippet-level prediction with a video classification loss. However, this formulation suffers from the discrepancy between classification and detection, resulting in inaccurate separation of foreground and background (F\&B) snippets. To alleviate this problem, we propose to explore the underlying structure among the snippets by resorting to unsupervised snippet clustering, rather than heavily relying on the video classification loss. 
Specifically, we propose a novel clustering-based F\&B separation algorithm. It comprises two core components: a snippet clustering component that groups the snippets into multiple latent clusters and a cluster classification component that further classifies the cluster as foreground or background. As there are no ground-truth labels to train these two components, we introduce a unified self-labeling mechanism based on optimal transport to produce high-quality pseudo-labels that match several plausible prior distributions.
This ensures that the cluster assignments of the snippets can be accurately associated with their F\&B labels, thereby boosting the F\&B separation.
We evaluate our method on three benchmarks: THUMOS14, ActivityNet v1.2 and v1.3. Our method achieves promising performance on all three benchmarks while being significantly more lightweight than previous methods. Code is available at ~\url{https://github.com/Qinying-Liu/CASE}

\end{abstract}

\section{Introduction}
\label{sec:intro}
Temporal action localization (TAL)~\cite{shou2016temporal,liu2020progressive,liu2023improve} is a task to localize the temporal boundaries of action instances and recognize their categories in videos. In recent years, numerous works put effort into the fully supervised manner and gain great achievements. Albeit successful, these methods require extensive manual frame-level annotations, which are expensive and time-consuming. 
Without the requirement of frame-level annotations, weakly-supervised TAL (WTAL) has received increasing attention, as it allows us to detect the action instances with only video-level action labels. 

\begin{figure*}[t]
	\centering
	\includegraphics[width=\linewidth]{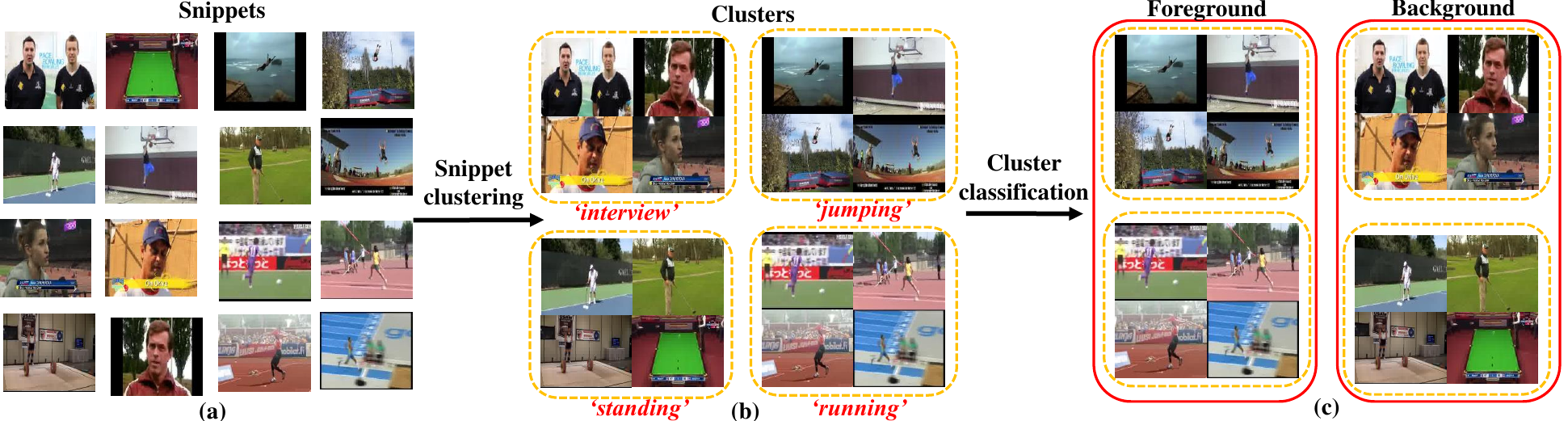}
	\centering
	\caption{Conceptual illustration of our clustering-based F\&B separation algorithm. In snippet clustering, we partition the snippets (or frames) into multiple clusters with explicit characteristics. In cluster classification, we classify the clusters as foreground or background.
	The above results are attained according to the predictions of our method. } 
	\label{fig:overview}
\end{figure*}

There has been a wide spectrum of WTAL methods developed in the literature~\cite{wang2017untrimmednets,Zhang_2021_cola,background_modeling,liu2023unleashing}. With only video-level labels, mainstream methods employ a localization-by-classification pipeline, which formulates WTAL as a video action classification problem to learn a temporal class activation sequence (T-CAS). For this pipeline, foreground (\textit{i.e.}, action) and background separation remains an open question since video-level labels do not provide any cue for background class. There are two types of existing approaches to solve it. The first type~\cite{wang2017untrimmednets,Zhang_2021_cola} is based on the multiple instance learning (MIL), which uses the T-CAS to select the most confident snippets for each action class. The second type~\cite{background_modeling,liu2019completeness} introduces an attention mechanism to learn class-agnostic foreground weights that indicate the probabilities of the snippets belonging to foreground. Despite recent progress, these methods typically rely on the video classification loss to guide the learning of the T-CAS or the attention weights. There is an inherent downside: the loss is easily minimized by the salient snippets~\cite{min2020adversarial} and fails to explore the distribution of the whole snippets, resulting in erroneous T-CAS or attention weights. This issue is rooted in the supervision gap between the classification and detection tasks. Recent studies~\cite{pardo2021refineloc,luo2020weakly} are devoted to producing snippet-level pseudo-labels to bridge the gap. However, the pseudo-labels are still derived from the unreliable T-CAS or attention weights. 

Deep clustering~\cite{chang2017deep}, which automatically partitions the samples into different groups, has been proven to be capable of revealing the intrinsic distribution of the samples in many label-scarce tasks~\cite{asano2019self,caron2020unsupervised,fini2021unified,li2022class,liu2022collaborating}. A natural issue arises: is it possible to adopt the clustering to capture the distribution of snippets?  Since clustering can be conducted in a self-supervised manner, it is immune to the video classification loss. This suggests a great potential of clustering for F\&B separation in WTAL. A brute-force solution  would be to group the snippets into two clusters, one for foreground and one for background. Whereas, we empirically find that it underperforms in practice (\textit{cf.}~\cref{exp:ablation}). We argue that the reason is that snippets, regardless of foreground or background, can differ dramatically in appearance (\textit{cf.}~\cref{fig:overview} (a)).
As a result, it may be difficult for a self-supervised model to group them accurately.  Fortunately, in real-world videos, there are common characteristics (\textit{e.g.}, "interview", "running") shared by a group of snippets (\textit{cf.}~\cref{fig:overview} (b)). Compared to learning two clusters for F\&B in the complex video content, it may be easier to explore the snippet clusters with clear and distinctive characteristics. This necessitates a clustering algorithm with multiple clusters. 
Furthermore, it can be observed that the characteristics of clusters are sometimes indicative cues for F\&B separation.
For example, we can confidently classify the "running" cluster to foreground and the "interview" cluster to background according to the cluster-level characteristics. 
Consequently, it is promising to further leverage the cluster-level representations to assist F\&B separation.

In light of the above discussion, we propose a novel \textbf{C}lustering-\textbf{A}ssisted F\&B \textbf{SE}paration (CASE) network. 
We begin by constructing a standard WTAL baseline that provides a primary estimation of F\&B snippets. 
We then introduce a clustering-based F\&B separation algorithm (\textit{cf.}~\cref{fig:overview}) to refine the F\&B separation. 
This algorithm is comprised of two main components: snippet clustering for dividing the snippets into multiple clusters, and cluster classification for classifying the clusters as foreground or background.
Considering that no ground-truth labels are available to train the components, we propose a unified self-labeling mechanism to generate high-quality pseudo-labels for them. Specifically, we formulate the label assignment in both components as a unified optimal transport problem, which allows us to flexibly impose several customized constraints on the distribution of pseudo-labels. 
After training these two components,  we can transform the cluster assignments of the snippets to their F\&B assignments, which can be used to refine the F\&B separation of the baseline. 

It is demonstrated that our method yields favorable performance while being much more lightweight compared to prior approaches.  In summary, our contributions are three-fold. 1) We propose a clustering-based F\&B separation algorithm for WTAL, which casts the problem of F\&B separation as a combination of snippet clustering and cluster classification. 2) We propose a unified self-labeling mechanism based on optimal transport to guide snippet clustering and cluster classification. 3) We conduct extensive experiments that demonstrate the effectiveness and efficiency of our method compared to existing approaches.

\section{Related Work} \label{sec:related_work}
\noindent\textbf{Deep clustering.} 
Current deep clustering approaches~\cite{yang2019deep,yang2020adversarial,dang2020multi} could be roughly divided into two categories. The first one iteratively computes the clustering assignment from the up-to-date model and supervises the network training processes by the estimated information~\cite{xie2016unsupervised,yang2016joint,chang2017deep,caron2018deep,chang2019deep,wu2019deep}. DeepCluster~\cite{xie2016unsupervised} is a typical method that iteratively groups the features and uses the subsequent assignments to update the deep network. 
The second one simultaneously learns feature representation and clustering assignment~\cite{haeusser2018associative,ji2019invariant,huang2020deep}, which has gained popularity in recent years. 
Asano~\textit{et al.}~\cite{asano2019self} propose to enforce a balanced label assignment constraint to avoid degenerate solution.
Caron~\textit{et al.}~\cite{caron2020unsupervised} use the algorithm in~\cite{asano2019self} to introduce a swapped mechanism that employs two random transformations of the same images to guide each other. In this work, we extend~\cite{asano2019self} from image classification to WTAL, and incorporate it with task-specific designs, \eg, imposing multiple sensible constraints on the distribution of pseudo-labels. 
It is worth noting that in the above methods, the number of clusters is typically set to the number of ground-truth (GT) classes so that clusters and GT classes can be mapped one-by-one during testing~\cite{van2020scan,caron2018deep}. There are attempts that utilize an extra over-clustering technique to learn a larger number of clusters than GT, which is believed to be conducive to representation learning~\cite{ji2019invariant,fini2021unified}. However, these methods commonly treat the technique as an auxiliary tool independent of their main task.
In contrast, we are committed to building an explicit correspondence between learned clusters and F\&B classes, thus unleashing the full potential of clustering in the WTAL task.   

\noindent\textbf{Weakly-supervised temporal action localization.}
Existing WTAL approaches can be categorized into four broad groups. The first group aims to improve feature discrimination ability. 
Various techniques, \eg, deep metric learning~\cite{min2020adversarial,Narayan20193CNetCC} and contrastive learning~\cite{Zhang_2021_cola,liexploring}, have been explored.
The second group seeks to discover complete action regions.~\cite{min2020adversarial,singh2017hide,Zhong2018StepbystepEO} hide some snippets to press the models in exploring more action regions, while~\cite{liu2019completeness,islam2021hybrid} use a multi-branch framework to discover complementary snippets. The third group is concerned with learning attention weights.~\cite{zhai2020two,nguyen2018weakly} design losses to regularize the values of the attention weights.~\cite{pardo2021refineloc,luo2020weakly} generate pseudo-labels for them. 
However, the pseudo-labels are derived from the primary predictions of snippets, which are still optimized using the video classification loss. 
The last group is the most closely related to ours, which introduces auxiliary classes in addition to the action classes.
~\cite{qu2021acm} introduces a video-level context class.~\cite{luo2021action,wang2021exploring,liu2019completeness} mine the action units or sub-actions shared across action categories.
Class-specific sub-action is explored in~\cite{huang2021two,huang2021modeling}. Recently,~\cite{li2022weakly} learns a set of visual concepts for fine-grained action localization. Our method is superior to these methods in three noticeable aspects. 1) These methods rely on video-level supervision to discover the auxiliary classes. Conversely, we develop the clusters in a self-supervised manner that is orthogonal to video-level supervision. 2) These methods devise multiple loss terms to regularize the auxiliary classes. In contrast, we introduce regularization into optimal transport, which can be resolved in a principled way. 3) Our method significantly outperforms these methods.

\section{Preliminaries and Baseline Setup} \label{sec:method_base}

In each training iteration, we first sample a mini-batch of $B$ videos. For each video, we have access only to its video-level label $\boldsymbol{Y} \in \mathbb{R}^G$,  where $G$ is the number of ground-truth action classes. By convention, we first sample a sequence of $T$ snippets from each video, and then extract snippet features with a pre-trained feature extractor for both RGB and optical-flow streams. For simplicity, only one stream is presented hereafter. As a result, we obtain a sequence of snippet features $\boldsymbol{F} \in \mathbb{R}^{T \times D}$. Here, $D$ is the channel dimension.

\begin{figure*}[t]
	\centering
	
	\includegraphics[width=0.95\linewidth]{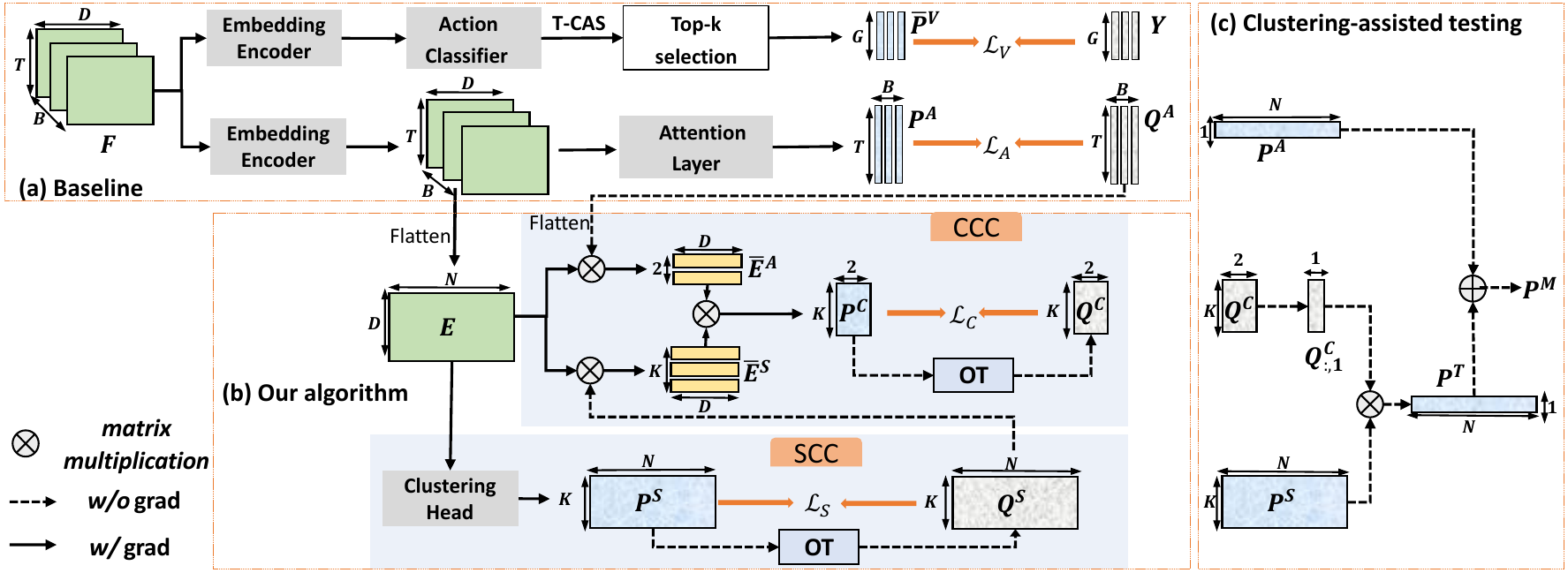}
	\centering
    
	\caption{Framework of our CASE. (a) depicts the baseline, which includes a video classification branch and an attention branch. (b) illustrates our proposed clustering-based F\&B separation algorithm, which comprises a snippet clustering component (SCC) and a cluster classification component (CCC). Both are trained using a unified self-labeling mechanism based on optimal transport (OT). (c) shows the clustering-assisted testing technique that utilizes the results of SCC and CCC to assist F\&B separation during inference. }
	
	\label{fig:framework}
\end{figure*}

For the baseline, following convention~\cite{lee2020background}, we use a two-branch framework consisting of a video classification branch and an attention branch, as shown in~\cref{fig:framework}(a). In the former branch, we first feed the input features $\boldsymbol{F}$ to an embedding encoder followed by an action classifier to get the temporal class activation sequence (T-CAS) $\boldsymbol{P^V} \in \mathbb{R}^{T \times G}$. In the latter branch, $\boldsymbol{F}$ is first passed through another embedding encoder to obtain the snippet embeddings, and the embeddings are then sent to an attention layer to extract one-dimension \textbf{A}ttention weights $\boldsymbol{P^A} \in \mathbb{R}^{T }$, which represent the foreground probabilities of the snippets.    

We apply the popular multiple instance learning (MIL) to train the video classification branch. Briefly (see Supplementary for details), 
we first calibrate T-CAS with the attention weights to restrain background snippets. Then we select the top-$k$ snippets for each class based on the activations to construct video-level scores $\boldsymbol{\bar{P}^V} \in \mathbb{R}^G$. Finally, we optimize a video classification loss with the known video labels $\boldsymbol{Y}$:
\begin{equation}
\begin{aligned} 
\mathcal{L}_{V} = \mathcal{L}_{CE} (\boldsymbol{\bar{P}^V},\boldsymbol{Y}).
\label{eq:video_loss}            
\end{aligned}
\end{equation}

To train the attention branch, we adopt the pseudo-label-based scheme proposed by~\cite{ma2021weakly} due to its conciseness and effectiveness. Specifically, we define foreground pseudo-labels $\boldsymbol{Q^A} \in \mathbb{R}^{T }$ as follows:  snippets appearing in the top-$k$ activations for the ground-truth video-level classes are positive, and the other snippets are negative. To improve the robustness of the model against label noise, we use the generalized binary cross-entropy loss~\cite{zhang2018generalized,ma2021weakly}:
\begin{equation}
\begin{aligned}
\mathcal{L}_{A} = &  \frac{1}{N_{\text{pos}}}  \sum_{t=1}^{T} \boldsymbol{Q^A}_{t} \frac{1 - {(\boldsymbol{P^A}_{t})}^\gamma}{\gamma}  + \\ & \frac{1}{N_{\text{neg}}} \sum_{t=1}^{T}  (1 - \boldsymbol{Q^A}_{t}) \frac{1 - {(1 - \boldsymbol{P^A}_{t})}^\gamma}{\gamma},
\label{eq:attention_loss}           
\end{aligned}
\end{equation}
where $\gamma \in (0, 1)$ controls the noise tolerance, and $N_{\text{pos}}$ and $N_{\text{neg}}$ represent the number of positive and negative snippets. 

\section{Our Method}
\label{sec:method_cfb_net} 

In this section, we present our clustering-based F\&B algorithm, as depicted in~\cref{fig:framework}(b), which is built upon the above baseline. We begin by providing an overview of our algorithm, which consists of two main components: snippet clustering component (SCC) and cluster classification component (CCC). Next, we introduce a unified self-labeling mechanism that we employ to provide pseudo-labels for both SCC and CCC. Lastly, we explain how SCC and CCC are used in the training and testing procedures.

\subsection{Overview} 
\paragraph{Snippet Clustering Component.} SCC is proposed to group snippets into latent clusters.  To enable joint learning of the attention layer and snippet clustering, we append SCC over the embeddings in the attention branch, as shown in~\cref{fig:framework}(b). For notation simplicity,  we henceforth term the total number of snippets as $N=BT$ for a batch of $B$ videos, and call the snippet embeddings $\boldsymbol{E} \in \mathbb{R}^{N \times D}$. 
We feed $\boldsymbol{E}$ into a clustering head composed of a linear classifier with $K$ classes (clusters), producing \textbf{S}nippet-level cluster assignment probabilities dubbed $\boldsymbol{P^S} \in \mathbb{R}^{N \times K}$. Inspired by self-supervised learning~\cite{pang2022unsupervised,qian2022unsupervised}, we set $K$ as a predefined parameter, which we find to be robust in practice. To train the clustering head, we first generate (soft) pseudo-labels $\boldsymbol{Q^S}  \in \mathbb{R}^{N \times K}$ for $\boldsymbol{P^S}$, which will be described in \cref{sec:method_sl}. Then, we minimize the following loss:
\begin{equation}
	\begin{aligned} 
		\mathcal{L}_S= \frac{1}{N} \sum_{n=1}^{N}  \mathcal{L}_{CE} (\boldsymbol{Q^S}_{n}, \boldsymbol{P^S}_{n}).
		\label{eq:SCC_loss}
	\end{aligned} 
\end{equation}

\paragraph{Cluster Classification Component.} CCC enforces each cluster to be classified into foreground or background by mapping the cluster prototypes to the F\&B prototypes, as shown in~\cref{fig:framework}(b). 
Specifically, based on the pseudo cluster assignments of snippets $\boldsymbol{Q^S} \in \mathbb{R}^{N \times K}$ obtained from SCC, we can compute the $k$-th cluster prototype over the snippet embeddings $\boldsymbol{E} \in \mathbb{R}^{N \times D}$:
\begin{equation}
	\begin{aligned} 
	\boldsymbol{\bar{E}^S}_k = \frac{\sum_{n=1}^{N} \boldsymbol{Q^S}_{n,k} \cdot \boldsymbol{E}_n }{\sum_{n=1}^{N} \boldsymbol{Q^S}_{n,K}}.
	\label{eq:cluster_proto}  
	\end{aligned}
\end{equation}
where $\boldsymbol{\bar{E}^S}_k \in \mathbb{R}^D$. In a similar vein, using the foreground pseudo-labels $\boldsymbol{Q^A} \in \mathbb{R}^{N }$ and background labels $1 - \boldsymbol{Q^A}$, we can calculate the F\&B prototypes $\boldsymbol{\bar{E}^A} \in \mathbb{R}^{2 \times D}$. $\boldsymbol{\bar{E}^A}_1$ and  $\boldsymbol{\bar{E}^A}_2$ correspond to foreground and background, respectively. 
Then, we compute \textbf{C}luster-level classification probabilities dubbed $\boldsymbol{P^C} \in \mathbb{R}^{K \times 2}$ by measuring the similarities between cluster prototypes and F\&B prototypes:
\begin{equation}
	\begin{aligned} 
	\boldsymbol{P^C}_{k,i} = \mathop{Softmax}\limits_{i} \big(\rho \cdot \operatorname{cos}(\boldsymbol{\bar{E}^S}_k, \boldsymbol{\bar{E}^A}_i)\big),
	\label{eq:pred}           
	\end{aligned}
\end{equation}
where $\operatorname{cos} (\cdot)$ indicates the cosine similarity function and $\rho$ is the temperature. $\boldsymbol{P^C}_{k,i}$ represents the probability that $k$-th cluster belongs to foreground ($i=1$) or background ($i=2$). To optimize the component, we generate (soft) labels $\boldsymbol{Q^C} $ for $\boldsymbol{P^C}$, which will be described in~\cref{sec:method_sl}. Accordingly, we will get a loss term:
\begin{equation}
	\begin{aligned} 
		\mathcal{L}_C= \frac{1}{K}\sum_{k=1}^{K} \mathcal{L}_{CE} (\boldsymbol{Q^C}_{k}, \boldsymbol{P^C}_{k}).
		\label{eq:CCC_loss}
	\end{aligned} 
\end{equation} 

Notably, although $\boldsymbol{Q^C}$ is computed from each mini-batch of data, we observe that it will quickly converge to a stable status near the one-hot form during training. This suggests that a global and clear correspondence between clusters and F\&B is established. 


\subsection{Self-Labeling via Unified Optimal-Transport} \label{sec:method_sl} 

This section explains the self-labeling mechanism that generates the labels $\boldsymbol{Q^S}$ for $\boldsymbol{P^S}$ in SCC and labels $\boldsymbol{Q^C}$ for $\boldsymbol{P^C}$ in CCC. First, we describe a basic labeling formulation shared in both SCC and CCC. This formulation converts the label assignment to an optimal transport problem while imposing constraints on the distribution of the labels. 
Then, we discuss the unique adaptations required for SCC and CCC individually. For SCC, we introduce a prior distribution for $\boldsymbol{Q^S}$ to avoid the uncertain label assignment issue observed in SCC. As for CCC, we leverage snippet-level F\&B labels to estimate the prior marginal distribution of $\boldsymbol{Q^C}$. These adaptations can be seamlessly integrated into the optimal-transport formulation, resulting in a unified solution that is easy to implement, as demonstrated in Alg.~\ref{alg:code}.

\paragraph{Basic formulation.} 
Regarding the generation of pseudo-labels $\boldsymbol{Q}$, a straightforward solution is to search for a reasonable $\boldsymbol{Q}$ that is close to the current model predictions $\boldsymbol{P}$, \eg, by applying $\operatorname{\arg \max}$ to $\boldsymbol{P}$. However, in our unsupervised setting, this way may lead to trivial solutions, \eg, all samples are assigned to only a class (\textit{cf.}~\cref{exp:ablation}). Instead, when searching for $\boldsymbol{Q}$, we propose to impose a constraint on the proportion of elements assigned to each class. Formally, we formulate this as an optimization problem:
\begin{equation}
\begin{aligned} 
\mathop{\min}_{\boldsymbol{Q} \in \Omega} \mathrm{E}(\boldsymbol P, \boldsymbol Q), 
\label{eq:objective}
\end{aligned} 
\end{equation}
where $\small{\mathrm{E}(\boldsymbol P, \boldsymbol Q)=-\sum_{n}^{N}\sum_{k}^{K} \boldsymbol{Q}_{n,k} \log \boldsymbol{P}_{n,k}}$ measures the distance between $\boldsymbol{Q}$ and $\boldsymbol{P}$, $N$ is the number of samples and $K$ is the number of classes. The constraint $\Omega$ is defined as:
\begin{equation}
	\begin{aligned} 
	\Omega = \{ \boldsymbol{Q} \in \mathbb{R}_{+}^{N \times K} | \boldsymbol{Q} \mathbf{1}^{K}= \boldsymbol{\alpha}, {\boldsymbol{Q}}^{\top} \mathbf{1}^N=\boldsymbol{\beta} \},
		\label{eq:contraint}
	\end{aligned} 
\end{equation}
where $\boldsymbol{\alpha}$ and $\boldsymbol{\beta}$ are the marginal distributions of $\boldsymbol{Q}$ onto its rows and columns, respectively.
We set $\boldsymbol{\alpha}=\mathbf{1}^N$ to ensure that $\boldsymbol{Q}$ is a probability matrix. 
$\boldsymbol{\beta} \in \mathbb{R}^{K}$ represents the proportion of elements belonging to each of the $K$ classes. When there is no prior knowledge,  equipartition~\cite{asano2019self,caron2020unsupervised} can serve as a general inductive bias and be utilized to set $\boldsymbol{\beta}$:
\begin{equation}
\begin{aligned} 
\boldsymbol{\beta} = \frac{N}{K} \mathbf{1}^{K}. 
\label{eq:equipartition_contraint}
\end{aligned} 
\end{equation}
This ensures that, on average, each class is assigned the same number of samples, thereby averting trivial solutions. 

It is noteworthy that~\cref{eq:objective} is an optimal transport problem, which is computationally expensive to solve. Following~\cite{cuturi2013sinkhorn}, an entropy term is introduced to it:
\begin{equation} 
	\begin{aligned} 
	\mathop{\min}_{\boldsymbol{Q} \in \Omega} \ \mathrm{E}(\boldsymbol P, \boldsymbol Q) - \frac{1}{\epsilon} \operatorname{H} (\boldsymbol{Q}) ,
	\label{eq:sk_objective}
	\end{aligned}
\end{equation}
where $\epsilon>0$ and $\operatorname{H}  (\boldsymbol{Q}) $ is the entropy of $\boldsymbol{Q}$. 
The advantage of the term is that~\cref{eq:sk_objective} can be efficiently solved by Sinkhorn-Knopp algorithm~\cite{cuturi2013sinkhorn}.

\paragraph{Prior distribution for SCC.} 
In~\cref{eq:sk_objective}, an entropy term is subtracted to make it tractable with affordable complexity. Nevertheless, maximizing the entropy can also lead to uncertain label assignments, where the samples are assigned to different classes with equal probability. In practice, the issue is pronounced in SCC but not in CCC. This may be because the former involves much more instances and classes, rendering the algorithm harder to converge (\textit{cf.}~\cref{fig:RSC}). 

To remedy the defect in SCC, we repurpose an early sequence-matching method~\cite{su2017order} to introduce a prior distribution for the pseudo-labels $\boldsymbol{Q^S}$ in SCC, denoted as $\boldsymbol{\hat{Q}^S} \in \mathbb{R}^{N \times K}$, which represents the probability of assigning $N$ snippets to $K$ clusters. A sensible prior distribution should encourage foreground snippets to have relatively high probabilities of belonging to foreground clusters, and the same is true for background.
To implement this, we first sort the snippets according to their foreground probabilities $\boldsymbol{P^A}$ in ascending order, and denote the resulting ranks of the $N$ snippets as ${rank} \in \mathbb{R}^N$. We then construct the prior distribution $\boldsymbol{\hat{Q}^S}$, such that the snippets with high ranks (\textit{i.e}, ${rank}$) are more likely to be assigned to the clusters with high foreground probabilities (\textit{i.e}, $\boldsymbol{Q^C}_{:,1}$) and vice versa.
Formally, $\boldsymbol{\hat{Q}^S}$ is defined as a Gaussian distribution:
\begin{equation}
	\begin{aligned} 
	\boldsymbol{\hat{Q}^S}_{n,k} = \frac{1}{\sigma\sqrt{2\pi}}\exp \big( -\frac{{| \frac{{rank}_n}{N} -\boldsymbol{Q^C}_{k,1}|}^2}{2\sigma^2}\big),
	\label{eq:Gaussian}
	\end{aligned}
\end{equation}
where ${rank}_n$ is the order of $n$-th snippet, and $\boldsymbol{Q^C}_{k,1}$ is the foreground probability of $k$-th cluster.
Finally, we replace~\cref{eq:sk_objective} with the following objective function:
\begin{equation}
	\begin{aligned} 
		\mathop{\min}_{\boldsymbol{Q^S} \in \Omega^S} \ \mathrm{E}(\boldsymbol{P^S}, \boldsymbol{Q^S})  + \frac{1}{\epsilon} \operatorname{KL} (\boldsymbol{Q^S}, \boldsymbol{\hat{Q}^S}),
		\label{eq:new_objective}
	\end{aligned}
\end{equation}
where $ \operatorname{KL}(\cdot)$ is the Kullback-Leibler divergence. 
By minimizing the KL term, we encourage the labels $\boldsymbol{Q^S}$ to be close to the prior distribution $\boldsymbol{\hat{Q}^S}$, which helps to avoid uncertain label assignments caused by the original entropy term.  Importantly,~\cref{eq:new_objective} can still be efficiently addressed by Sinkhorn-Knopp algorithm. 
For detailed derivation, we refer to Supplementary.

\paragraph{Prior marginal distribution for CCC.} 
Although equipartition (\ie,~\cref{eq:equipartition_contraint}) is a common prior in traditional clustering, it is not suitable for enforcing equipartition on the marginal distribution of the cluster-level F\&B labels $\boldsymbol{Q^C}$, namely $\boldsymbol{\beta^C} \in \mathbb{R}^{2}$. This is due to the fact that $\boldsymbol{\beta^C}$ represents the proportions of clusters assigned to F\&B, which are not always balanced.  However, since SCC enforces equipartition on the snippet level, each cluster contains a similar number of snippets. Consequently, the proportions of F\&B clusters are expected to be close to the proportions of F\&B snippets. To this end, instead of using~\cref{eq:equipartition_contraint}, we estimate $\boldsymbol{\beta^C}$ empirically based on the distribution of the snippet-level foreground labels $\boldsymbol{Q^A}$:
\begin{equation}
	\begin{aligned} 
		\boldsymbol{\beta^C} = [\frac{1}{N}\sum_{n=1}^N \boldsymbol{Q^A}_n, \ \frac{1}{N}\sum_{n=1}^N (1 - \boldsymbol{Q^A}_n)].
	\end{aligned} 
	\label{eq:beta}
\end{equation}

\begin{algorithm}[t]

\caption{\small Pseudo-code of labeling procedures of SCC and CCC in Pytorch-like style.
	}
	\label{alg:code}
	\definecolor{codeblue}{rgb}{0.25,0.5,0.5}
	\definecolor{codekw}{rgb}{0.85, 0.18, 0.50}
	\lstset{
		backgroundcolor=\color{white},
		basicstyle=\fontsize{7.3pt}{7.3pt}\ttfamily\selectfont,
		columns=fullflexible,
		breaklines=true,
		captionpos=b,
		numbers=left,
		xleftmargin=1em,
		commentstyle=\fontsize{7.3pt}{7.3pt}\color{codeblue},
		keywordstyle=\fontsize{7.3pt}{7.3pt}\color{codekw},
		escapechar=\&
	}
	\begin{lstlisting}[language=python]
# L_S: logit output of clustering head  (NxK)
# Q_hat_S: prior distribution in SCC (NxK)
# L_C: logit output of cluster classification (Kx2)
# Beta_C: prior marginal distribution in CCC (2)

# generating labels Q_S for SCC
Q_S = SK(L_S, Q_hat=Q_hat_S) # (NxK)
# generating labels Q_C for CCC
Q_C = SK(L_C, Beta=Beta_C) # (Kx2)

# resolve the optimal-transport problem by iterative Sinkhorn-Knopp (SK) algorithm
def SK(L, Q_hat=None, Beta=None, n_iter=3):
    N, K = L.size()
    # uniform distribution is the default when prior distribution is not given
    Q_hat = 1/K if Q_hat is None else Q_hat
    Beta = 1/K if Beta is None else Beta
    Q = exp(L / eps)
    Q = (Q * Q_hat).T 
    Q /= Q.sum()
    for _ in range(n_iter):
        Q /= Q.sum(dim=1)  
        Q = Q * Beta 
        Q /= Q.sum(dim=0) 
        Q /= N
    return Q.t() * N  
	\end{lstlisting}

\end{algorithm}

\subsection{Training and Testing}

\paragraph{Joint training.} \label{sec:method_train_and_test}  
We train all components together in an end-to-end fashion.  
The overall objective is written as:  
\begin{equation} 
	\begin{aligned} 
	\mathcal{L} = (\mathcal{L}_V  + \mathcal{L}_A) + \lambda_S \mathcal{L}_S + \lambda_C \mathcal{L}_C,
	\label{eq:overall_objective}	
	\end{aligned}
\end{equation}
where $\lambda_{S}$ and $\lambda_{C}$ represent the loss weights. As the baseline and our proposed algorithm share the same embedding encoder in the attention branch, the joint training also facilitates the training of the baseline model.

\paragraph{Clustering-assisted testing.} \label{sec:method_tfl}
In the inference period, using the cluster-level foreground probabilities $\boldsymbol{Q^C}_{:,1}$, we can transform the snippet-level cluster assignments $\boldsymbol{P^S}$ to snippet-level foreground probabilities dubbed $\boldsymbol{P^T}$ based on~\textit{law of total probability}: $\boldsymbol{P^T} = \boldsymbol{P^S}\boldsymbol{Q^C}_{:,1}$, as depicted in~\cref{fig:framework} (c). Considering that $\boldsymbol{Q^C}$ is stable during training, we simply use the $\boldsymbol{Q^C}$ from the last training iteration for inference. Moreover, as verified in~\Cref{table:effect_of_cat}, the transformed foreground probability $\boldsymbol{P^T}$ is complementary to the foreground probability $\boldsymbol{P^A}$ from the attention layer. Hence, we fuse $\boldsymbol{P^A}$ and $\boldsymbol{P^T}$ by convex combination: $\boldsymbol{P^M} = 0.5\boldsymbol{P^A} + 0.5\boldsymbol{P^T}$.  The combined probability $\boldsymbol{P^M}$ is then used to help localize action instances during inference.

\section{Experiments}

\subsection{Datasets and Evaluation Metric}

\textbf{THUMOS14}~\cite{THUMOS14} contains videos with $20$ classes. We use the $200$ videos in validation set for training and the $213$ videos in testing set for evaluation. 
\textbf{ActivityNet}~\cite{caba2015activitynet} has two release versions, \textit{i.e.}, \textbf{ActivityNet v1.3} and \textbf{ActivityNet v1.2}.
\textbf{ActivityNet v1.3}~\cite{caba2015activitynet} covers 200 action categories with 10, 024 and 4, 926 videos in the training and validation sets, respectively. \textbf{ActivityNet v1.2} is a subset of ActivityNet v1.3, and covers 100 action categories with 4, 819 and 2, 383 videos in the training and validation sets, respectively 
We follow the standard evaluation protocol by reporting mean Average Precision (mAP) under various temporal intersection over union (tIoU) thresholds.
We refer to \textbf{Supplementary} for more experimental details about architecture, setup, baseline,~\textit{etc}.

\begin{table}[t]
	\begin{center}
	\setlength{\tabcolsep}{2.5pt}
		\resizebox{\columnwidth}{!}{
			\begin{tabular}{lcccccccccc}
				\hline
				\multirow{2}{*}{Method}  &\multirow{2}{*}{MACs} & \multirow{2}{*}{Params} &  \multicolumn{4}{c}{mAP @ IoU} & \multirow{2}{*}{\makecell{AVG\\0.1:0.5}} & \multirow{2}{*}{AVG} \\
				\cline{4-7}
				& & & 0.1 & 0.3 & 0.5 & 0.7 &  & \\
				\hline
				W-TALC~\cite{paul2018w} & -& -& 55.2 & 40.1 & 22.8 & 7.6 & 39.8 & - \\
				BaS-Net~\cite{lee2020background} & 38.60 & 26.26 & 58.2 & 44.6 & 27.0 & 10.4 & 43.6 & 35.3 \\
				TSCN~\cite{zhai2020two} &- & -& 63.4 & 47.8 & 28.7 & 10.2 & 47.0 & 37.8 \\
				ACM-Net~\cite{qu2021acm}  & 9.48 & 12.63 & 68.9 & 55.0 & 34.6 & 10.8 & 53.2 & 42.6  \\
				CoLA~\cite{Zhang_2021_cola} & 9.47 & 12.62 & 66.2 & 51.5 & 32.2 & 13.1 & 50.3 & 40.9 \\
				UGCT~\cite{yang2021uncertainty} &- & -& 69. & 55.5 & 35.9 & 11.4 & 54.0 & 43.6 \\
				CO$_2$-Net~\cite{hong2021cross} &20.88 & 34.13 & 70.1 & 54.5 & 38.3  & 13.4 & 54.4 & 44.6 \\
				FTCL~\cite{gao2022fine} &9.45 & 14.73 & 69.6 & 55.2 & 35.6 & 12.2 & 53.8 & 43.6\\
				RSKP~\cite{huang2022weakly} & 6.90 & 4.20 & 71.3 & 55.8 & 38.2  & 12.5 & 55.6 & 45.1 \\
				ASM-Loc~\cite{he2022asm} & 13.77 & 46.21 & 71.2 & 57.1 & 36.6 & 13.4 & 55.4 & 45.1 \\
                DELU~\cite{chen2022dual} &20.88 & 34.13 & 71.5 & 56.5 & 40.5 & 15.5 & 56.5 & \textbf{46.4} \\
				\rowcolor{gray!10} \textbf{CASE}  & \textbf{1.60} & \textbf{2.13}  & 72.3 & 59.2 & 37.7 & 13.7 & \textbf{57.1} & 46.2 \\
				\hline
		\end{tabular}}
	\end{center}
\vspace{-3mm}
\caption{Comparison on THUMOS14 in terms of mAP (\%), MACs (G), and Params (M). AVG(0.1:0.5) and AVG represent the average mAP under IoU thresholds of 0.1:0.5 and 0.1:0.7. MACs are computed from a video with 750 snippets.}
\label{table:THUMOS14}
\end{table}

\begin{table}[t]
\setlength{\tabcolsep}{4pt}
	     \begin{center}
  		\resizebox{\columnwidth}{!}{
  			\begin{tabular}{lcccc}
  			\hline
  			 & Our baseline & CO$_2$-Net~\cite{hong2021cross} & AICL~\cite{li2023actionness} &Zhou~\textit{et al.}~\cite{zhou2023improving}  \\
  			\hline 
  			Base        & 42.1 & 44.6 & 46.4 & 48.3 \\ 
  			Base + CASE & 46.2 & 46.6 & 48.0 & 49.2 \\ 
  			\hline
  		\end{tabular}}
	    \end{center}
\vspace{-3mm}
\caption{Compatibility with different baselines. }
\label{table:generalization_analysis}
\end{table}

\subsection{Comparison with SOTA Methods}

In~\Cref{table:THUMOS14}, we compare our CASE with SOTA WTAL methods on THUMOS14. Besides mAP, we also report the model complexity in terms of multi-accumulative operations (MACs), and the number of trainable parameters (Params). 
We can observe that CASE achieves comparable performance to the recent SOTA method DELU~\cite{chen2022dual} and evidently outperforms other approaches. Notably, CASE is significantly more lightweight than the competitors, with MACs and Params less than 1/10 of those of DELU. This is because our method does not require a heavy cross-modal consensus module~\cite{chen2022dual,hong2021cross} or multi-step proposal refinement~\cite{he2022asm}. We can increase the complexity of CASE by combining it with previous methods. The results are presented in Table~\ref{table:generalization_analysis}. We can observe that our CASE consistently and evidently boosts the performances of all the baselines, highlighting its strong compatibility with these baselines.

Additionally, we compare our CASE with previous methods on ActivityNet v1.2\&v1.3 in Table~\ref{table:activity}. As can be seen, CASE achieves impressive improvements over the previous methods on both datasets.

\begin{table}[!t]
\begin{subtable} {0.49\columnwidth}
\begin{center}
\setlength{\tabcolsep}{2pt}
		\begin{center}
			\resizebox{\columnwidth}{!}{
				\begin{tabular}{lcccc}
					\hline
					\multirow{2}{*}{Method} & \multicolumn{4}{c}{mAP @ IoU} \\
					\cline{2-5}
					& 0.5 & 0.75 & 0.95 & AVG \\
					\hline
					BaS-Net  \cite{lee2020background} & 38.5 & 24.2 & 5.6 & 24.3 \\
					RPN~\cite{huang2020relational} &	37.6 &	23.9 &	5.4 & 23.3 \\
					EM-MIL~\cite{luo2020weakly}  &37.4 & - &- & 20.3 \\
					TSCN~\cite{zhai2020two} & 37.6 & 23.7 & 5.7 & 23.6 \\
					WUM  \cite{lee2021weakly} & 41.2 & 25.6 & 6.0 & 25.9 \\
					CoLA \cite{Zhang_2021_cola} & 42.7 & 25.7 & 5.8 & 26.1 \\
					ASL \cite{ma2021weakly}  & 40.2 & -  & - & 25.8 \\
					D2-Net \cite{narayan2021d2} & 42.3 & 25.5 & 5.8 & 26.0 \\
					ACGNet \cite{yang2021acgnet} & 41.8 & 26.0 & 5.9 & 26.1 \\
                    DELU \cite{chen2022dual} & 44.2 & 26.7 & 5.4 & 26.9   \\
					\rowcolor{gray!10} \textbf{CASE} & 43.8 & 27.2 & 6.7 & \textbf{27.9} \\
					\hline
			\end{tabular}}
		\end{center}

\caption{ActivityNet v1.2}
\label{table:activity1.2}
\end{center}
\end{subtable} 
\hfill
\begin{subtable} {0.49\columnwidth}
\begin{center}
	\setlength{\tabcolsep}{2pt}
	\resizebox{\columnwidth}{!}{
				\begin{tabular}{lcccc}
					\hline
					\multirow{2}{*}{Method} & \multicolumn{4}{c}{mAP @ IoU} \\
					\cline{2-5}
					& 0.5 & 0.75 & 0.95 & AVG \\
					\hline
					BaS-Net  \cite{lee2020background} & 34.5 & 22.5 & 4.9 & 22.2\\
					TSCN  \cite{zhai2020two} & 35.3 & 21.4 & 5.3 & 21.7\\
					MSA  \cite{huang2021modeling} & 36.5 & 22.8 & 6.0 & 22.9\\
					ACM-Net \cite{qu2021acm} & 40.1 & 24.2 & 6.2 & 24.6 \\
					UGCT \cite{yang2021uncertainty} & 39.1 & 22.4 & 5.8 & 23.8 \\
					AUMN  \cite{luo2021action} & 38.3 & 23.5 & 5.2 & 23.5\\
					FAC-Net \cite{huang2021foreground} & 37.6 & 24.2 & 6.0 & 24.0 \\
					FTCL \cite{gao2022fine} & 40.0 & 24.3 & 6.4 & 24.8 \\
					RSKP \cite{huang2022weakly} & 40.6 & 24.6 & 5.9 & 25.0 \\
					ASM-Loc \cite{he2022asm} & 41.0 & 24.9 & 6.2 & 25.1 \\
					\rowcolor{gray!10} \textbf{CASE} & 43.2 & 26.2 & 6.7 & \textbf{26.8} \\
					\hline
			\end{tabular}}

\caption{ActivityNet v1.3}
\label{table:activity1.3}
\end{center}
\end{subtable} 

\caption{Results on ActivityNet v1.2\&v1.3. AVG indicates the average mAP at IoU thresholds 0.5:0.05:0.95.}
\label{table:activity}

\end{table}

\subsection{Ablation Study} \label{exp:ablation}

\noindent\textbf{Contribution of core components.} 
Our method contains two core components: snippet clustering component (SCC) and cluster classification component (CCC), which are successively stacked on the baseline model and jointly trained with it. \Cref{table:each_component} quantifies the contributions of each component.
When compared to the baseline (line \#1), both components contribute to the performance.
Specifically, when the clustering-assisted testing technique is not applied, SCC and CCC yield gains of 1.1\% (line \#2) and 0.7\% (line \#3), respectively, indicating that the joint training with SCC and CCC indeed improves the performance of the baseline. The reason is that through deliberated clustering of snippets, the embedding space is shaped as well-structured, as illustrated in~\cref{fig:tsne}, thereby facilitating the learning of the baseline. 
After training SCC and CCC, we leverage the clustering-assisted testing to further make use of SCC and CCC during testing, which brings about a substantial promotion of 2.3\% (line \#4). In a nutshell, SCC and CCC are beneficial in both the training and testing stages.

\begin{table}[!t]

	\begin{center}
	\setlength{\tabcolsep}{7pt}
		\resizebox{\columnwidth}{!}{
		\begin{tabular}{clcccl}
		\hline
		\multirow{2}{*}{\#}  & \multirow{2}{*}{Method}  & \multicolumn{4}{c}{mAP @ IoU (\%)} \\
        \cline{3-6}
		& & 0.3& 0.5 & 0.7 & AVG   \\
		\hline
		1 & Baseline & 53.8 & 31.9 & 11.9 & 42.1   \\
		\hline
		2 & + SCC & 55.3 & 33.7 & 12.4 & 43.2$_{\textcolor{blue}{+1.1}}$  \\
		3 & + SCC + CCC & 56.1 & 34.9 & 12.8 & 43.9$_{\textcolor{blue}{+0.7}}$  \\
		\rowcolor{gray!10}  4 & + SCC + CCC (T) & 59.2 & 37.7 & 13.7 & 46.2$_{\textcolor{blue}{+2.3}}$  \\
		\hline
		\end{tabular}
		}
	\end{center}

\captionsetup{skip=0pt}
\caption{Ablation study of core components. "(T)" indicates that clustering-assisted testing is applied.}
\label{table:each_component}
\end{table}

\noindent\textbf{Necessity of converting label assignment to optimal transport (OT).} 
We propose an OT-based labeling strategy that generates labels by solving an OT problem: the objective~\cref {eq:objective} subject to a constraint~\cref{eq:contraint} on the proportion of each class. To assess the necessity of our design, we use the labeling strategy in SCC as an example and compare it with alternative labeling strategies. The results are summarized in~\Cref{table:pseudo-labeling}. The terms "Hard" and "Soft" indicate whether the labels are one-hot or soft, respectively. 
The strategy without OT, denoted as "\textit{w/o} OT", refers to solving~\cref {eq:objective} without~\cref{eq:contraint}, which is no longer an OT problem and can be solved immediately. In particular, in the case of "\textit{w/o} OT+Soft", the labels would be nearly identical to the model predictions, resulting in almost no learning. As a consequence, it shows poor performance. On the other hand, "\textit{w/o} OT+Hard" involves applying $\operatorname{\arg \max}$ to the model predictions, but it also performs poorly. To understand why, we calculate the entropy of the proportions of clusters, namely $\operatorname{H}(\boldsymbol{\bar{P}^S})$, where $\boldsymbol{\bar{P}^S} = \frac{1}{N} \sum_{n}^N \boldsymbol{P^S}_n$. In~\Cref{table:pseudo-labeling}, we observe that "\textit{w/o} OT+Hard" results in a low $\operatorname{H}(\boldsymbol{\bar{P}^S})$ close to zero, indicating a trivial solution where most snippets are assigned to only a few clusters. In contrast, our proposed "\textit{w/} OT" greatly alleviates this issue and achieves much better performance. Notably, "\textit{w/} OT+Hard" performs worse than "\textit{w/} OT+Soft", which we attribute to the aggressive nature of obtaining hard labels. Overall, the results confirm the efficacy of our design.

\begin{table}[!t]
\setlength{\tabcolsep}{15pt}
	\begin{center}
  		\resizebox{\columnwidth}{!}{
  			\begin{tabular}{lccccc}
  			\hline
  			\multirow{2}{*}{Metric} & \multicolumn{2}{c}{\textit{w/o} OT} & \multicolumn{2}{c}{\textit{w/} OT} \\
                \cline{2-5}
  			 & Soft & Hard & Soft & Hard\\
  			\hline  
  			mAP  & 41.2 & 40.8 & 46.2 & 44.7 \\
                $\operatorname{H}(\boldsymbol{\bar{P}^S})$ & 2.77 & 0.01 & 2.76 & 2.75 \\
  			\hline
  		\end{tabular}}
	\end{center}

\captionsetup{skip=0pt}
\caption{Ablation study of OT-based labeling.}
\label{table:pseudo-labeling}
\end{table}

\begin{figure}[t]
\centering
 \resizebox{1.04\columnwidth}{!}{
     \begin{minipage}{0.48\linewidth}{
	\includegraphics[width=1.05\linewidth]{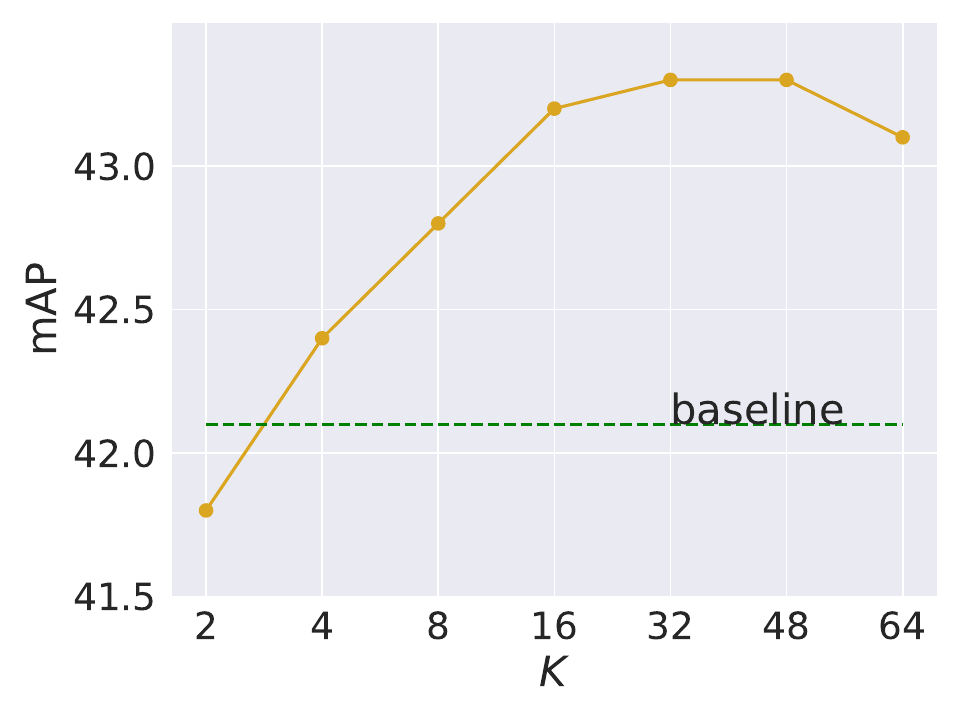}

	\caption{Ablation study of the number of clusters $K$. } 
  \label{fig:KC}
    }
    \end{minipage}
    \hspace{0.7mm}
   \begin{minipage}{0.48\linewidth}{
	\includegraphics[width=1.05\linewidth]{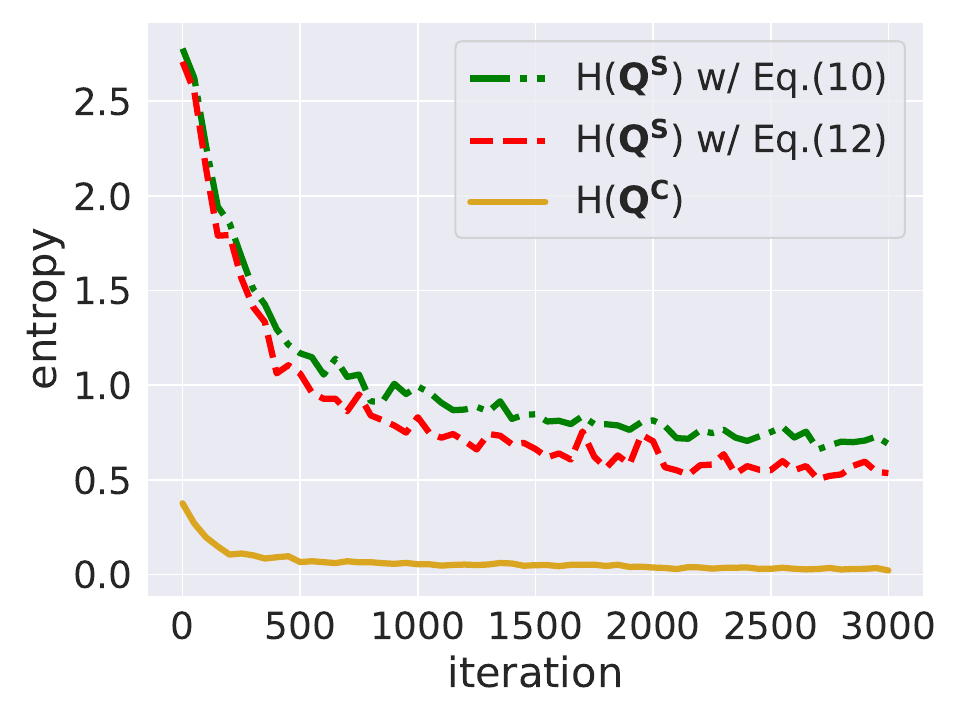}
	\centering

	\caption{The evolution of $H(\boldsymbol{Q^C})$ and $H(\boldsymbol{Q^S})$. } 
  	\label{fig:RSC}
   }
    \end{minipage}
}
\end{figure}

\noindent\textbf{Analysis of the number of clusters $K$.} We cluster snippets into multiple clusters ($K > 2$), even though only F\&B separation is required. To verify the correctness of our design, we compare the performances under different numbers of $K$ in~\cref{fig:KC}.
As can be seen, a small $K$ results in inferior performance, with $K=2$ even causing a performance decline relative to baseline. The reason may be that the clustering results deviate too much from the true distribution of F\&B snippets. Hence, clustering into multiple clusters is necessary. Besides, the performance becomes stable ($\pm$0.2\%) and robust to $K$ as long as enough clusters are set ($K>16$), making it easy to tune an appropriate $K$ in practice. These findings are in accordance with self-supervised learning~\cite{qian2022unsupervised,pang2022unsupervised,wang2021unsupervised}. 

\begin{table}[!t]
\setlength{\tabcolsep}{15pt}
\begin{center}
  	\resizebox{\columnwidth}{!}{
  			\begin{tabular}{p{0.02cm}<{\centering}p{4cm}p{1cm}<{\centering}}
  			  \hline
  			    \#	& Method  & mAP \\
  				\hline 
  				\rowcolor{gray!10} 1 & our CASE & 46.2 \\ 
  				2 & using~\cref{eq:sk_objective} in SCC  & 45.4 \\ 
  				3 & using~\cref{eq:equipartition_contraint} in CCC  & 44.0 \\ 
  				\hline
  		\end{tabular}}
\end{center}

\captionsetup{skip=0pt}
\caption{Analysis of unique techniques in SCC and CCC.  }

\label{table:effect_of_SL}
\end{table}

\noindent\textbf{Analysis of snippet clustering component (SCC). } 
In SCC, we enforce pseudo-labels $\boldsymbol{Q^S}$ to match the prior distribution $\boldsymbol{\hat{Q}^S}$ 
by transitioning the objective of OT problem from the original objective~\cref{eq:sk_objective} to the current objective~\cref{eq:new_objective}. 
To evaluate the impact of this modification, in~\Cref{table:effect_of_SL}, we compare the results of our method (line \#1) against those obtained using the original~\cref{eq:sk_objective} (line \# 2). The results show that our modified objective outperforms the original one, indicating that matching the prior distribution is beneficial.
To further examine this phenomenon, in~\cref{fig:RSC}, we plot the evolution of the entropy of $\boldsymbol{Q^S}$, $\operatorname{H}(\boldsymbol{Q^S})$, using current~\cref{eq:new_objective} or original~\cref{eq:sk_objective}.  A large $\operatorname{H}(\boldsymbol{Q^S})$ is typically undesired as it indicates that the assignment is uncertain. 
Additionally, we report the entropy of $\boldsymbol{Q^C}$ in CCC, $\operatorname{H}(\boldsymbol{Q^C})$, as an indication, though a direct comparison is not entirely fair. The results reveal that $\operatorname{H}(\boldsymbol{Q^C})$ converges rapidly to a small value, indicating that optimization can easily yield a solution close to a one-hot form in CCC. Conversely, $\operatorname{H}(\boldsymbol{Q^S})$ converges slowly and remains high, suggesting uncertain assignments in SCC. However, this issue is mitigated when using our proposed~\cref{eq:new_objective}. 

\noindent\textbf{Analysis of cluster classification component (CCC). }  In CCC, we empirically estimate the prior marginal distribution of pseudo-labels $\boldsymbol{Q^C}$ based on~\cref{eq:beta}, as opposed to using equipartition~\cref{eq:equipartition_contraint}. To evaluate the impact of this design choice,  we present the results of using equipartition~\cref{eq:equipartition_contraint} in line \#3 of \Cref{table:effect_of_SL}. The clear superiority of our method over the use of equipartition highlights the importance of using a data-driven approach to estimate the prior distribution of pseudo-labels in CCC.

\begin{table}[!t]
\setlength{\tabcolsep}{20pt}
	\begin{center}
  		\resizebox{\columnwidth}{!}{
  			\begin{tabular}{cccc}
  			  \hline
  			    Method & $\boldsymbol{P^A}$ & $\boldsymbol{P^T}$ & $\boldsymbol{P^M}$ \\
  				\hline 
                    mAP & 43.9 & 44.1 & 46.2 \\
  				\hline
  		\end{tabular}}
 	\end{center}

\captionsetup{skip=0pt}
\caption{Analysis of the clustering-assisted testing.  }

\label{table:effect_of_cat}
\end{table}

\begin{figure}[t]
	\centering
	\includegraphics[width=1\linewidth]{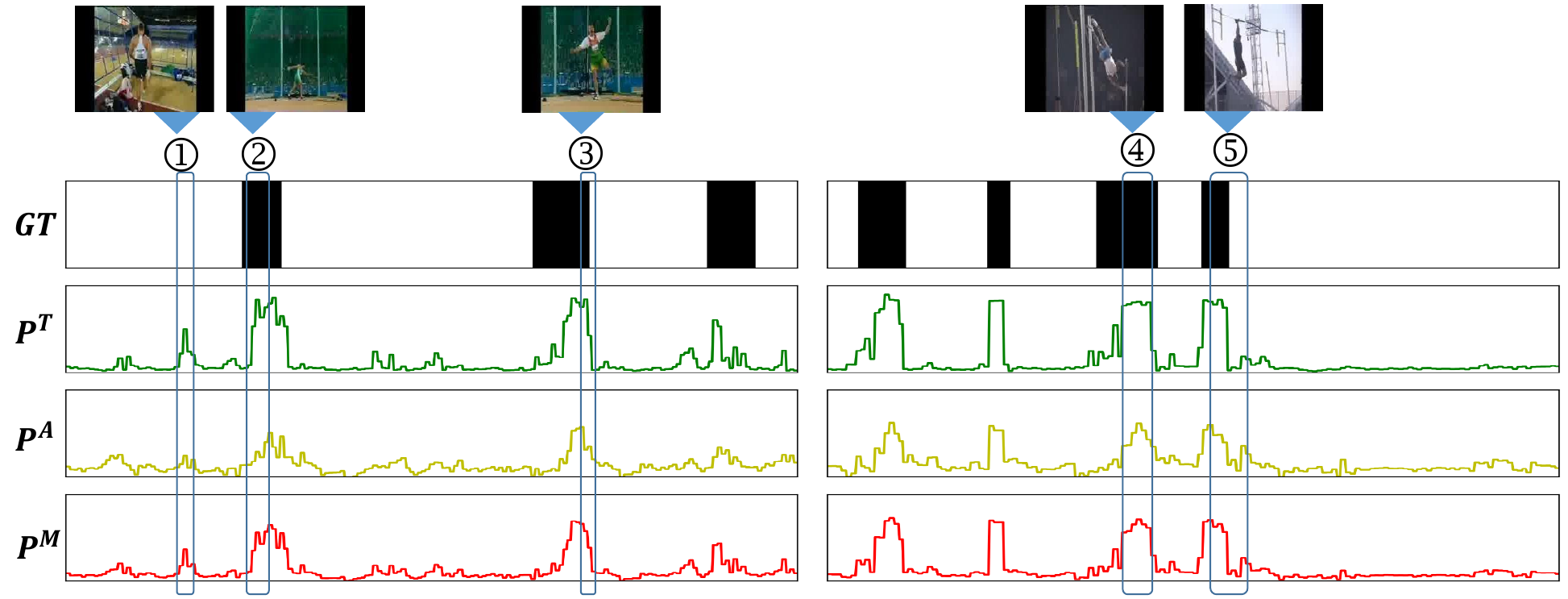}
	\centering
	\caption{Qualitative results of two videos on THUMOS14. We show $\boldsymbol{P^T}$, $\boldsymbol{P^A}$ and $\boldsymbol{P^M}$, and ground-truth (GT).}
	\label{fig:vis}
\end{figure}

\noindent\textbf{Analysis of clustering-assisted testing.}  We propose to fuse the foreground probabilities $\boldsymbol{P^A}$ from the baseline model and the foreground probabilities $\boldsymbol{P^T}$ transformed from the cluster assignments, resulting in the fused one $\boldsymbol{P^M}$.
In~\Cref{table:effect_of_cat}, we compare the performances of $\boldsymbol{P^A}$, $\boldsymbol{P^T}$, and $\boldsymbol{P^M}$. It can be seen that the fused one $\boldsymbol{P^M}$ outperforms both $\boldsymbol{P^A}$ and $\boldsymbol{P^T}$, which well demonstrates that $\boldsymbol{P^A}$ and $\boldsymbol{P^T}$ are complementary to each other. 
This is further confirmed by the qualitative results exemplified in~\cref{fig:vis}. Concretely, it delivers the following inspiring observations: 1) Compared to $\boldsymbol{P^A}$, $\boldsymbol{P^T}$ tends to activate more action-related regions (\textit{e.g.}, \ding{173}, \ding{175}) and clearer action boundaries (\textit{e.g.}, \ding{174}, \ding{176}). This is in line with our motivation for using $\boldsymbol{P^T}$, which is designed to be more independent of the video classification loss, capturing a more comprehensive distribution of snippets rather than being biased by discriminative regions. 2) $\boldsymbol{P^T}$ is not always superior to $\boldsymbol{P^A}$. When optimizing $\boldsymbol{P^T}$, CASE regards the snippets as independent samples and fails to make use of the video-level labels, resulting in irrelevant action activation (\textit{e.g.}, \ding{172}). These observations further support the complementary nature of $\boldsymbol{P^A}$ and $\boldsymbol{P^T}$.

\begin{figure}[t]
	\centering
	\includegraphics[width=1\linewidth]{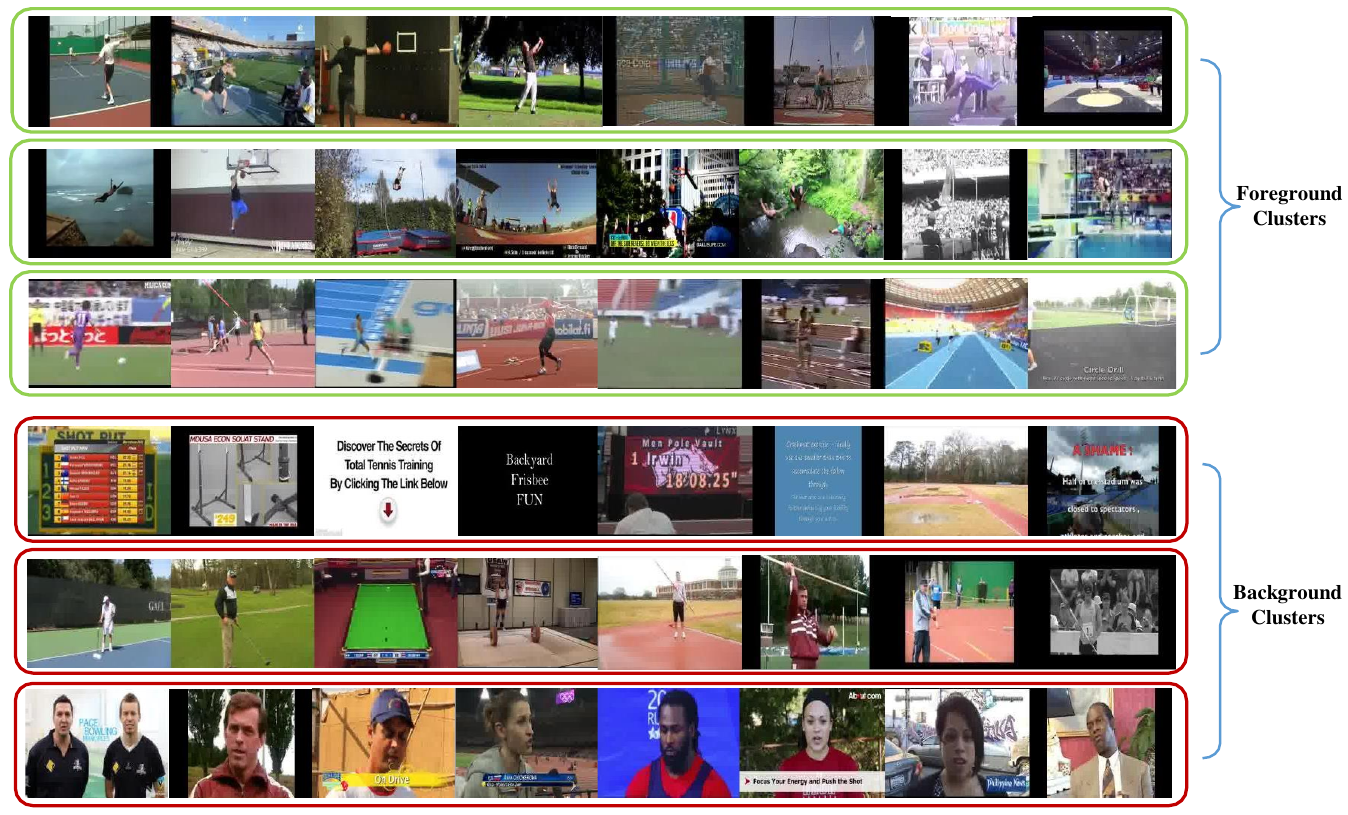}
	\centering
	\caption{Qualitative results of  snippet clustering and cluster classification. We show three clusters belonging to foreground at the top and there clusters belonging to background at the bottom.}
	
	\label{fig:example}
\end{figure}

\begin{figure}[t]
	\centering
	\includegraphics[width=\linewidth]{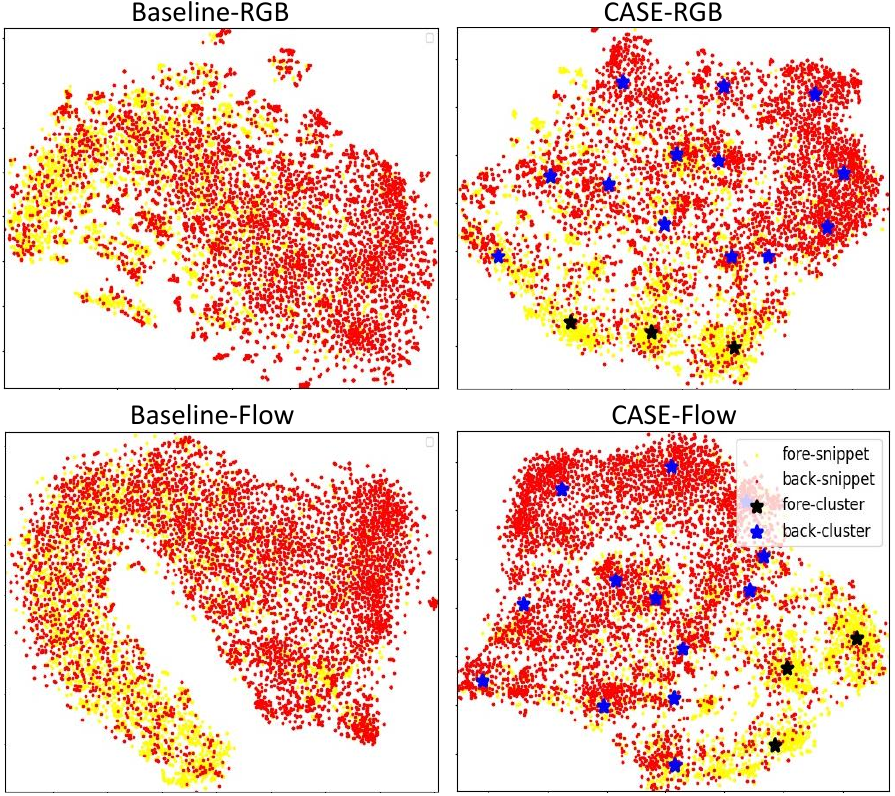}
	\centering
	\caption{TSNE visualization of $10, 000$ snippet embeddings and $16$ cluster prototypes.  
    "-RGB" and "-Flow" indicate RGB stream and optical-flow stream, respectively.
	The yellow, red, black, and blue marks represent foreground snippets, background snippets, foreground cluster prototypes, and background cluster prototypes, respectively.} 
	\label{fig:tsne}
\end{figure}

\noindent\textbf{Visualization results.} \label{sec:vis}
To gain more insight, we illustrate qualitative results of snippet clustering and cluster classification in~\cref{fig:example}. It can be seen that: 1) Our method  effectively identifies the snippet groups with common characteristics, such as "swinging arms" for the 1st row, "standing" for the 5th row. 2) Our method is accurate in classifying clusters into F\&B classes. For instance, some foreground frames in 1st\&3rd rows and background frames in 5th row are visually similar, yet our method can still separate them into different clusters with correct F\&B labels.
Overall, these results vividly substantiate the efficacy of our method in effectively grouping and classifying snippets. 

Furthermore, we visualize the snippet embeddings and cluster prototypes  in~\cref{fig:tsne} and make the following observations: 1) Each cluster prototype is surrounded by a couple of snippets, and the cluster prototypes are distinguishable between foreground and background. 
2) The embedding space learned by our method exhibits a clearer boundary between foreground and background classes compared to the baseline. These results clearly demonstrate that 1) the learned clusters are highly representative of the snippets; 2) the embedding space in our method is well-shaped and captures a comprehensive distribution of the snippets.

\section{Conclusion and Limitation} \label{sec:conclusion}

In this work, we propose a WTAL framework named CASE, which leverages snippet clustering to improve F\&B separation. Specifically, CASE comprises a snippet clustering component that partitions snippets into multiple clusters, followed by a cluster classification component that identifies the F\&B clusters. To optimize these components, we employ a unified self-labeling strategy based on optimal transport. Our extensive analysis demonstrates the effectiveness and efficiency of CASE. One limitation of our method is the requirement for a WTAL baseline to provide semantic-level reference of F\&B classes to classify the clusters into F\&B. A more self-contained clustering-based framework is our future work.

\section{Acknowledgements}
This work is supported by the National Natural Science Foundation of China under Grant 62176246 and Grant 61836008. This work is also supported by Anhui Province Key Research and Development
Plan (202304a05020045), Anhui Provincial Natural Science Foundation 2208085UD17 and the Fundamental Research Funds for the Central Universities (WK3490000006).

{\small
\bibliographystyle{ieee_fullname}
\bibliography{egbib}
}

\clearpage
\appendix
{
  \hypersetup{linkcolor=black}
  \tableofcontents
}

\section{Notation} \label{A:notation}  
In~\Cref{Tab.notation_defination}, we summarize the key notations used in our paper for easy reference.

\section{Architecture} \label{A:arch}  
\Cref{table:arch} presents the details of the architecture of our proposed CASE. 
As can be seen, the architecture is simple and lightweight, consisting of only a few 1D convolutional layers with a kernel size of 1 and linear layers, leading to the high efficiency of our method.

\begin{table}[!h]
\centering
\footnotesize
\setlength{\tabcolsep}{5pt}
\resizebox{\columnwidth}{!}{
\begin{tabular}{lcll}
\hline
 & \textbf{Notation} & \textbf{Shape}  & \textbf{Description} \\
\hline
\multirow{3}{*}{Baseline} & $\boldsymbol{P^A}$ &  $N$ & foreground probability \\
& $\boldsymbol{Q^A}$ &  $N$ & pseudo-labels of $\boldsymbol{P^A}$ \\
& $\boldsymbol{Z}$ &  $N \times D$ & snippet embedding \\
\hline
\multirow{3}{*}{SCC}& $\boldsymbol{P^S}$ & $N \times K$ & cluster assignment probability \\
& $\boldsymbol{Q^S}$ & $N \times K$ & pseudo-labels of $\boldsymbol{P^S}$ \\
& $\boldsymbol{\hat{Q}^S}$ & $N \times K$ & prior distribution of $\boldsymbol{Q^S}$ \\
\hline
\multirow{3}{*}{CCC}& $\boldsymbol{P^C}$ & $K \times 2$ & cluster classification probability \\
& $\boldsymbol{Q^C}$ & $K \times 2$ & pseudo-labels of $\boldsymbol{P^C}$ \\
& $\boldsymbol{\beta^C}$ & $2$ & prior marginal distribution of $\boldsymbol{Q^C}$ \\
\hline
\multirow{2}{*}{Testing}& $\boldsymbol{P^T}$ & $N \times K$ & transformed foreground probability \\
& $\boldsymbol{P^M}$ & $N \times K$ & fused foreground probability \\
\hline
\end{tabular}
}
\caption{Key notations in this paper. }
\label{Tab.notation_defination}
\end{table}

\begin{table}[!h]
\centering
\small
\resizebox{\columnwidth}{!}{
 \begin{tabular}{l|c|cccc|c}
\hline
 component & layer & kernel & stride& dim & act & output size\\
\hline
& \multicolumn{6}{c}{Embedding Encoder} \\
\cline{2-7}
\multirow{7}{*}{Baseline} & $\operatorname{Conv1d}$ & 1 &  1 &  512 & $\operatorname{Relu}$ & 512$\times T$\\
\cline{2-7}
 & \multicolumn{6}{c}{Action Classifier } \\
\cline{2-7}
 & $\operatorname{Conv1d}$ & 1 &  1 &  $G$ & $\operatorname{Softmax}$ & $G \times T$\\
\cline{2-7}
 & \multicolumn{6}{c}{Embedding Encoder} \\
\cline{2-7}
 & $\operatorname{Conv1d}$ & 1 &  1 &  512 & $\operatorname{Relu}$ & 512$\times T$\\
\cline{2-7}
  & \multicolumn{6}{c}{Attention layer} \\
\cline{2-7}
 & $\operatorname{Conv1d}$ & 1 &  1 &  1 & $\operatorname{Sigmoid}$ & $ 1 \times T$\\
\hline
\multirow{2}{*}{Our Algorithm}  & \multicolumn{6}{c}{Clustering Head} \\
\cline{2-7}
 & $\operatorname{Linear}$ & 1 &  1 &  $K$ & $\operatorname{Softmax}$ & $ K \times T$\\
\hline
\end{tabular}
}
\caption{The detailed architecture of CASE, where the RGB stream and optical flow stream share the same structure. }
\label{table:arch}
\end{table}

\section{Additional Ablation Experiments} \label{A:Exp_on_Abl}  

\subsection{Ablation on multiple datasets} To show the effectiveness of our method in various scenarios, we perform a component-wise ablation study for the snippet clustering component (SCC) and the cluster classification component (CCC) on THUMOS14, ActivityNet v1.2 and v1.3. The corresponding results are provided in~\Cref{table:each_component}. We observe consistent trends across all datasets, indicating the robustness and effectiveness of our approach.

\subsection{Analysis of baseline model } We carry out several ablation experiments to analyze the components of the baseline. The results are illustrated in~\Cref{table:baseline}. It can be seen that the attention branch largely increases the performance, demonstrating the significance of class-agnostic F\&B separation. Additionally, we find that the use of the generalized binary cross-entropy loss yields better results than the traditional binary cross-entropy loss, proving that enhancing the label noise tolerance is advantageous.

\subsection{Analysis of ranking indices ${rank}$ } \label{A:Analysis_rank}
\begin{figure}[!t]
  	\begin{center}
  		\includegraphics[width=0.7\linewidth]{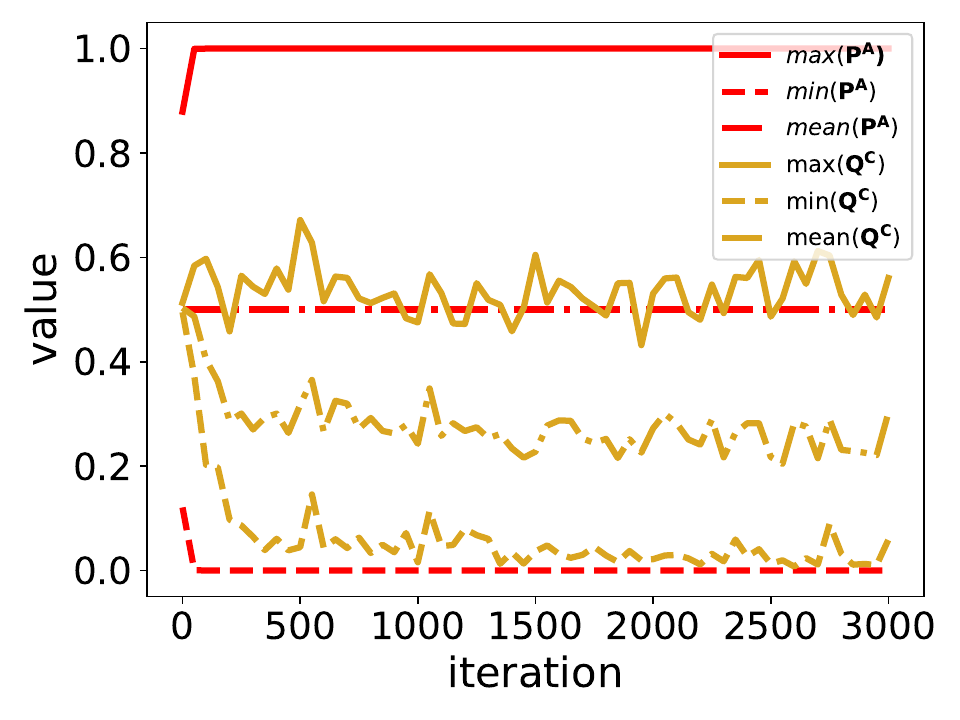}
  		\caption{The maximum, average, and minimum values of $\boldsymbol{P^A}$ and $\boldsymbol{Q^C}$ of each iteration during training. }
  		\label{fig:scr}
  	\end{center}
 \end{figure}

In SCC, we use the distance between the normalized ranking indices of the snippets~${rank}/N$ and the cluster-level pseudo-labels $\boldsymbol{Q^C}$ to compute a 2D gaussian distribution. In principle, ${rank}/N$ can be replaced by $\boldsymbol{P^A}$. However, we experimentally find that the performance of using $\boldsymbol{P^A}$ is inferior to that of using ${rank}/N$ (\textit{i.e.}, $45.1$ for $\boldsymbol{P^A}$~\textit{vs.} $46.2$ for ${rank}/N$ on average mAP). To explain it, we show the statistics (~\textit{i.e.}, maximum, average and minimum) of $\boldsymbol{P^A}$ and $\boldsymbol{Q^C}$ in~\cref{fig:scr}. The statistics are computed over each batch (\textit{i.e.}, iteration). Notably, the maximum, average, and minimum of ${rank}/N$ are always $\frac{1}{N} \simeq 0$, $0.5+ 0.5\frac{1}{N}\simeq 0.5$ and $1$, respectively.   
As can be seen, compared with $\boldsymbol{P^A}$, ${rank}/N$ is more comparable to $\boldsymbol{Q^C}$. For example,  both the average of $\boldsymbol{Q^C}$ and the average of ${rank}/N$ are around $0.5$ and are evidently larger than the average of $\boldsymbol{P^A}$.  This observation confirms the validity of our approach.

\begin{table}[!t]
	\begin{center}
	\setlength{\tabcolsep}{5pt}
		\resizebox{\columnwidth}{!}{
		\begin{tabular}{lccc}
		\hline
		& THUMOS14 &ActivityNet v1.2 &ActivityNet v1.3   \\
		\hline
		Baseline & 42.1 & 25.6 & 24.7  \\
		\hline
		+ SCC & 43.2 & 26.5 & 25.4 \\
		+ SCC + CCC & 43.9 & 27.0 & 25.7  \\
		 \rowcolor{gray!10} + SCC + CCC (T) & 46.2 & 27.9 & 26.8  \\
		\hline
		\end{tabular}
		}
	\end{center}
	\caption{Component-wise ablation study on THUMOS14, ActivityNet v1.2 and v1.3. "(T)" indicates that the clustering-assisted testing technique is appiled.}
	\label{table:each_component}
\end{table}

\begin{table}[!t]
\setlength{\tabcolsep}{20pt}
\begin{center}
\resizebox{\columnwidth}{!}{
\begin{tabular}{ccc|c}
\hline
VTB & ATB &  GBCE & mAP \\
\hline 
\checkmark & &  & 32.0  \\ 
\checkmark & \checkmark & & 41.7   \\ 
\checkmark & \checkmark  & \checkmark & 42.1  \\ 
\hline
\end{tabular}}
\end{center}
\caption{Ablation study on the baseline. VTB, ATB, and GBCE indicate video classification branch, attention branch, and generalized binary cross-entropy loss, respectively. Notably, if GBCE is not used, we use the traditional binary cross-entropy loss to train the ATB.}
\label{table:baseline}
\end{table}

\begin{figure*}[t]
	\centering
	\includegraphics[width=\linewidth]{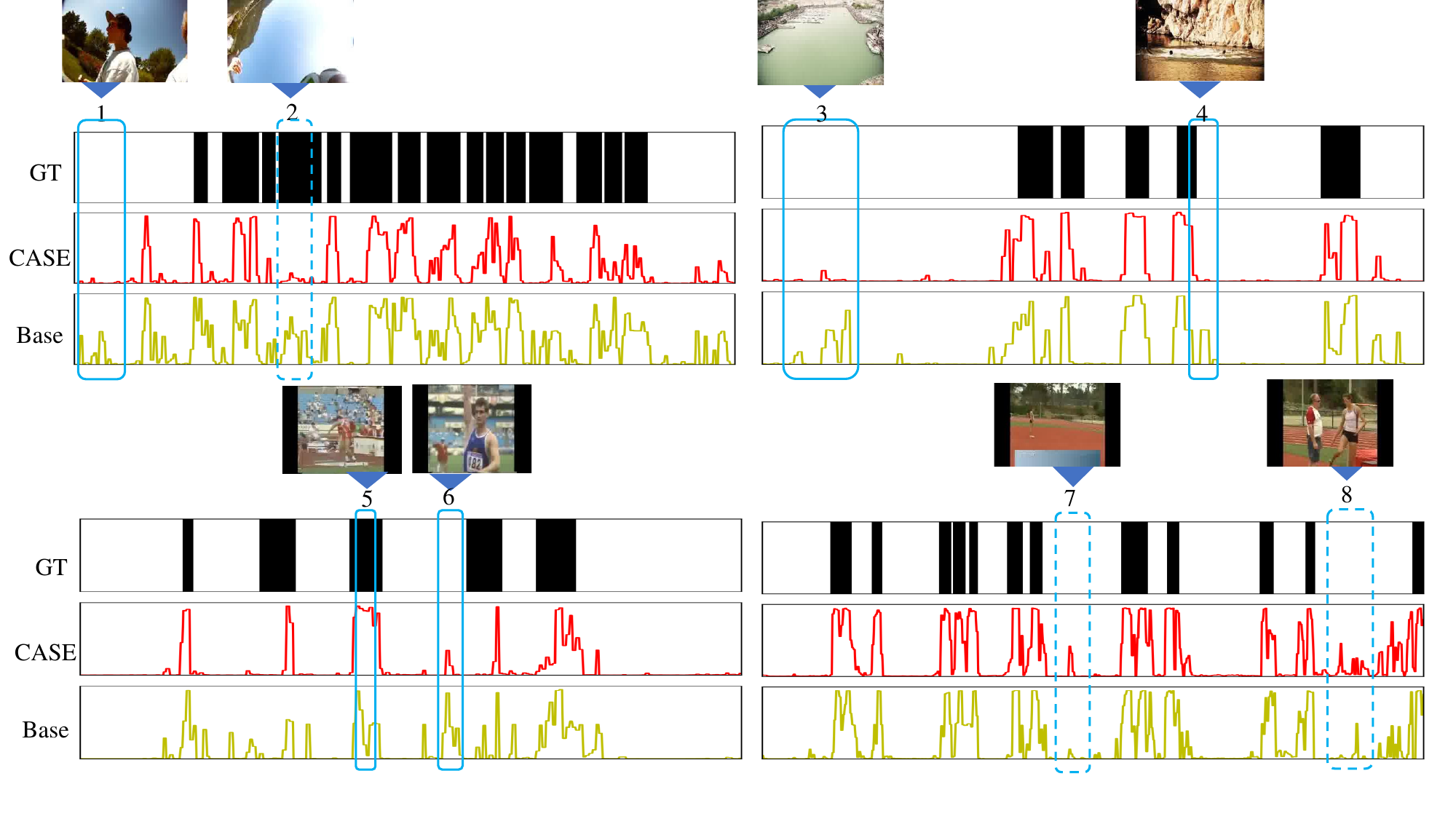}
	\centering
	\caption{Comparison between our CASE and the baseline. The solid and dashed boxes represent the regions where CASE outperforms and underperforms the baseline, respectively. } 
	\label{fig:case_vs_base}
\end{figure*}

\section{Additional Visualizations} \label{A:Vis}

\subsection{Comparison to baseline} 
In~\cref{fig:case_vs_base}, four visualized examples are provided to illustrate the differences between the F\&B separation results of CASE and that of baseline. It can be observed that: 1) CASE is advantageous to capture fine-grained patterns of snippets that are helpful to distinguish different snippets (see the solid boxes). For instance, in the region of '4', which is near the boundary of a 'diving' action instance, the foreground snippets and the background snippets are visually similar. However, CASE can accurately classify these snippets into correct F\&B classes, whereas the baseline cannot, showing that CASE can capture the underlying fine-grained structure of the snippets. 2) CASE performs worse than the baseline in some 'suspicious' regions (see the dashed boxes). To name a few, in the region of '8', an athlete raises her leg, causing CASE to mistake the region for an action instance. This mistake may be avoided by the baseline model because the video-level labels used to train the baseline can offer instructive information for the potential action types within the videos.  

\begin{figure*}[t]
	\centering
	\includegraphics[width=\linewidth]{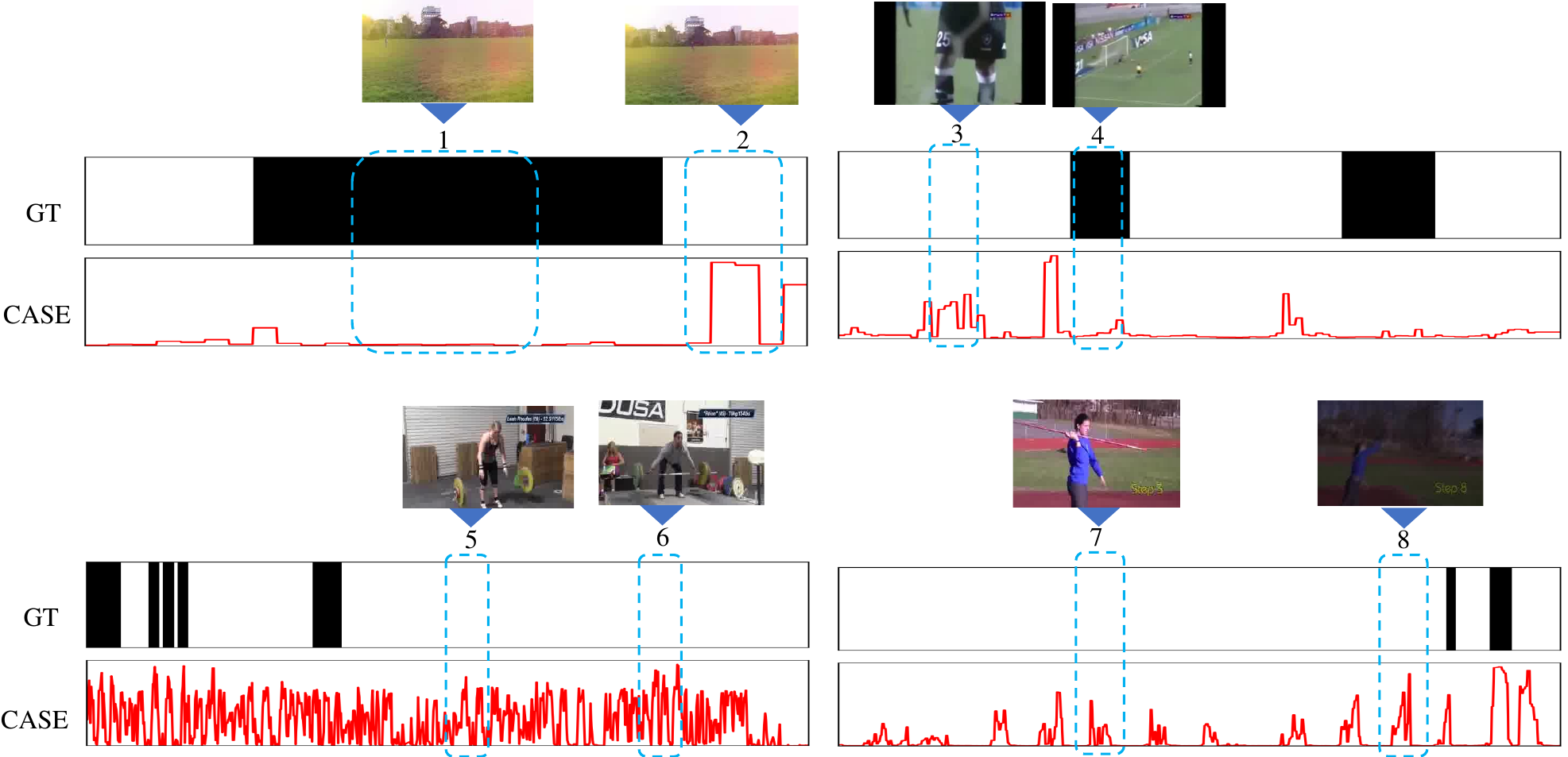}
	\centering
	\caption{Samples of failure cases. The dashed boxes represent the regions with wrong predictions. } 
	\label{fig:failure_cases}
\end{figure*}

\subsection{Failure cases} We showcase some examples of failure cases of our method in~\cref{fig:failure_cases}. From the figure, we conjecture the reason that accounts for the failure cases are: 1) low quality of images,~\textit{e.g.}, '1' and '8'; 2) indistinguishable body motions,~\textit{e.g.}, '3' and '7'; 3) small objects,~\textit{e.g.}, '2' and '4'; 4) incorrect annotation,~\textit{e.g.}, '5' and '6'.  These challenging cases represent future directions for our work.

\section{Additional Discussion on Related Work}  \label{A:Diss}
In our main paper, we extensively discussed the differences between our method and previous deep clustering and WTAL methods in the Related Work section. In this section, we would like to provide additional insights on other related methods.

In the snippet clustering component (SCC), we draw inspiration from the early sequence-matching method~\cite{su2017order} to construct a prior distribution for the pseudo-labels of cluster assignments of snippets. However, our method differs significantly from~\cite{su2017order} in both purpose and solution. Specifically,~\cite{su2017order} aims to measure the distance between two sequences by matching the frames of one sequence with the frames of another sequence with similar temporal positions. Consequently, it constructs a prior distribution for the mapping between the frames in different sequences based on their temporal locations. In contrast, we aim to disambiguate the assignments between snippets and clusters by enforcing the snippets with high foreground/background probabilities to be assigned to the clusters with high foreground/background probabilities. To achieve this, we construct the prior distribution for the cluster assignments of the snippets based on the distance between the foreground probabilities of snippets and the foreground probabilities of clusters. Besides, to better suit our approach, we rank the snippets according to their foreground probabilities, resulting in ranking indices that are more comparable with the foreground probabilities of the clusters. This approach allows us to better match the snippets to the appropriate clusters, as demonstrated in~\cref{A:Analysis_rank}.

Furthermore, our method is somewhat related to context-based methods~\cite{qu2021acm,liu2021weakly}. 
Previous context-based methods~\cite{qu2021acm,liu2021weakly} typically regard the context as a special type of background. That is, they divide foreground and background snippets into \textit{three latent groups}: action, context, and normal background. This approach provides a more detailed description of the background distribution. Our method extends this approach by dividing snippets into \textit{multiple latent groups}, which allows for a more comprehensive description of both the foreground and background distributions. The visualized results (\textit{e.g.}, 5th row of Fig.6 in the main paper) reveal that some of the learned clusters are very close to the concept of context. From this view,  our model already has some contextual modeling capabilities.

\section{Implementation Details} \label{A:Implementation}  
\subsection{Baseline model}  \label{A:Baseline}
Here we present more details about the multiple instance learning (MIL) used in the baseline. Specifically, we first calibrate T-CAS $\boldsymbol{P^V} \in \mathbb{R}^{T \times G}$ with the attention weights $\boldsymbol{P^A} \in \mathbb{R}^{T}$ to highlight foreground snippets and suppress background snippets, resulting in the calibrated T-CAS (dubbed $\boldsymbol{\hat{P}^V} \in \mathbb{R}^{T \times G}$). It can be implemented in multiple ways. 
Here following~\cite{ma2021weakly,qu2021acm},  we fuse the scores by weighted summation, $\boldsymbol{\hat{P}^V} = \omega \boldsymbol{P^V} + (1 - \omega)  \boldsymbol{P^A} $. $\omega$ is a predefined weight. Thereafter, we select $K$ snippets from each video for each class based on $\boldsymbol{\hat{P}^V}$: 
\begin{equation}
\begin{aligned}\small 
 \Gamma_{c}=\arg\max_{\begin{subarray}{}\Gamma\subset \{1,..,T\} \\\quad |\Gamma|=K \end{subarray} } \sum_{\tau \in \Gamma}{\boldsymbol{\hat{P}^V}_{\tau,c}},
\label{eq:topk_selection}        
\end{aligned}
\end{equation}
where $K$ is a hyper-parameter. Temporal pooling is applied to the selected snippets in $ \Gamma_{c}$ to build video-level class prediction $\boldsymbol{\bar{P}} \in \mathbb{R}^{G}$:
\begin{equation}
\begin{aligned}\small 
 {\boldsymbol{\bar{P}}_{c}= \mathop{Softmax}_c (\frac{1}{K} \sum_{\tau \in \Gamma_{c}}{\boldsymbol{P^V}_{\tau,c}}}) 
\label{eq:video_prediction}.       
\end{aligned}
\end{equation}
Finally, $\boldsymbol{\bar{P}}$ is used to compute a video classification loss, as shown in the main paper.

\subsection{Co-labeling} \label{sec:method_tscl} 
In our framework, there are several procedures of pseudo-labeling that can be summarized with a unified formulation as $\boldsymbol{Q} = \Psi(\boldsymbol{P})$.
Here $\boldsymbol{P}$ is the prediction of the model, $\Psi$ is the function of generating pseudo-labels, $\boldsymbol{Q}$ is the pseudo-labels.
To improve the quality of the pseudo-labels, following~\cite{zhai2020two}, we propose to apply the two-stream co-labeling (TSCL) strategy, which is model-agnostic and naturally compatible with our method. That is, we aggregate the predictions of RGB and optical-flow streams to generate the modality-sharing pseudo-labels, \textit{i.e.}, $\boldsymbol{Q} = \Psi(0.5 \boldsymbol{P^{\text{RGB}}} + 0.5 \boldsymbol{P^{\text{Flow}}})$. 
To be specific, for $\boldsymbol{Q^C}$, we fuse the cluster assignments of RGB stream (dubbed $\boldsymbol{P^{C,\text{RGB}}}$) and that of Flow stream (dubbed $\boldsymbol{P^{C,\text{Flow}}}$) by:
\begin{equation} 
	\begin{aligned}\small 
\boldsymbol{P^C} = 0.5  \boldsymbol{P^{C,\text{RGB}}} +0.5 \boldsymbol{P^{C,\text{Flow}}}.
		\label{eq:p_0}
	\end{aligned} 
\end{equation}
Then the pseudo-labels $\boldsymbol{Q^C}$ is generated by:
\begin{equation}
	\begin{aligned}\small 
		\mathop{\min}_{\boldsymbol{Q^C} \in \Omega^C} \ \langle \boldsymbol{Q^C}, -\operatorname{log} \boldsymbol{P^C} \rangle
		\label{eq:q_0}.
	\end{aligned} 
\end{equation}
As for $\boldsymbol{Q^R}$, the prediction of cluster classifier of RGB stream (dubbed $\boldsymbol{P^{R,\text{RGB}}}$) and that of Flow stream (dubbed $\boldsymbol{P^{R,\text{Flow}}}$) are combined as follows
\begin{equation} 
	\begin{aligned}\small 
\boldsymbol{P^R} = 0.5  \boldsymbol{P^{R,\text{RGB}}} +0.5 \boldsymbol{P^{R,\text{Flow}}}.
		\label{eq:p_1}
	\end{aligned} 
\end{equation}
Then the pseudo-labels $\boldsymbol{Q^R}$ is generated by:
\begin{equation}
	\begin{aligned}\small 
		\mathop{\min}_{\boldsymbol{Q^R} \in \Omega^R} \ \langle \boldsymbol{Q^R}, -\operatorname{log} \boldsymbol{P^R} \rangle.
		\label{eq:q_1}
	\end{aligned} 
\end{equation}
Moreover, the top-$K$ selection used in~\cref{eq:topk_selection} can be regarded as a procedure of defining the F\&B snippets. Hence, we utilize the TSCL to improve the quality of the top-$K$ selection. Specifically, we fuse the calibrated T-CAS of RGB stream (dubbed $\boldsymbol{\hat{P}^{V, \text{RGB}}}$) and that of optical-flow stream (dubbed $\boldsymbol{\hat{P}^{V, \text{Flow}}}$) as follows:
\begin{equation} 
	\begin{aligned}\small 
\boldsymbol{\hat{P}^{V}} = 0.5  \boldsymbol{\hat{P}^{V, \text{RGB}}} +0.5 \boldsymbol{\hat{P}^{V, \text{Flow}}}.
		\label{eq:p_2}
	\end{aligned} 
\end{equation}
Then $\boldsymbol{\hat{P}^{V}}$ is used for top-$K$ selection.
Notably, the results of the top-$K$ selection also influences the definition of $\boldsymbol{Q^A}$.
we use~\cref{eq:topk_selection} to determine the foreground and background snippets.
, which can influence the learning of both the video classification module and attention module. 

In Table~\ref{table:TSCL}, we present an evaluation of the effect of the two-stream co-labeling strategy on both the baseline model and our clustering-based F\&B algorithm. It can be seen that the TSCL is important to the baseline, boosting its performance from $38.3\%$ to $42.1\%$. However, the additional use of the TSCL in our algorithm results in only a small improvement compared to not using the TSCL in our algorithm (from $45.6\%$ to $46.2\%$). This suggests that the main reason for the performance improvement of our algorithm over the baseline model is our proposed clustering-based approach, rather than the two-stream co-labeling strategy.

\begin{table}[!h]
\setlength{\tabcolsep}{15pt}
		\begin{center}
  		\resizebox{\columnwidth}{!}{
  			\begin{tabular}{l|c}
  			\hline
  			Method & mAP \\
  			\hline 
                Baseline \textit{w/o} TSCL & 38.3  \\ 
  			Baseline \textit{w/} TSCL & 42.1 \\ 
  			\hline
                Baseline \textit{w/} TSCL + Our algorithm \textit{w/o} TSCL & 45.6 \\ 
  			Baseline \textit{w/} TSCL + Our algorithm \textit{w/} TSCL & 46.2 \\ 
  			\hline
  		\end{tabular}}
	    \end{center}
  	\caption{Ablation study of two-stream co-labeling (TSCL).}
  	\label{table:TSCL}
\end{table}

\subsection{Training details} \label{A:train_details}
TVL1~\cite{zach2007duality} is applied to extract optical-flow stream from RGB stream in advance. Each stream is divided into $16$-frame snippets. 
Following convention, we employ the I3D~\cite{carreira2017quo} network pre-trained on Kinetics-$400$~\cite{carreira2017quo} to extract snippet-level features from each stream, where the channel dimension $D$ is $1024$.
The number of sampled snippets $T$ is set to $750$ for THUMOS14 and $50$ for ActivityNet v1.2 and v1.3. 
Both streams share the same structure but have separate parameters.
The embedding encoders are comprised of a temporal convolution layer with $512$ channels and a $\operatorname{ReLU}$ layer. The action classifier consists of a $\operatorname{FC}$ layer and a $\operatorname{Softmax}$ layer. The clustering head is composed of a linear cosine classifier~\cite{gidaris2018dynamic} with a temperature of $10$ and a $\operatorname{Softmax}$ layer.   
The attention layer consists of a $\operatorname{FC}$ layer and a $\operatorname{Sigmoid}$ layer.
We set the classes $K$ of the clustering head to $16$ for THUMOS14 and $64$ for ActivityNet v1.2 and v1.3. 
Following previous methods~\cite{qu2021acm,ma2021weakly}, the $k$ for top-$k$ selection is set to $T//8$ in THUMOS14 and $T//2$ in ActivityNet v1.2 and v1.3, while the batch size $B$ is set to $16$, the $\gamma$ is set to $0.7$ and the $\omega$ is set to $0.25$ for all datasets.  
Following~\cite{caron2020unsupervised}, the $\epsilon$ is set to $20$. The temperature $\rho$ is set to $10$. The standard deviation $\sigma$ is set to $10$. 
The loss weights are set as $\lambda_S=1, \lambda_C=0.3$ for all datasets.
We utilize Adam optimizer with a learning rate of $10^{-4}$ for all datasets. We run each experiment three times and report their mean accuracy for reliability.
The model implemented by Pytorch is trained on a Nvidia 1080Ti GPU.

\subsection{Testing details} \label{A:test_details}
During inference, the video-level scores and snippet-level scores (\ie, T-CAS) of both the RGB stream and optical-flow stream are fused by averaging. Then, a threshold is applied to the video-level scores to determine the action categories. For the selected action class, a threshold strategy is applied to the T-CAS, as done in~\cite{qu2021acm,gao2022fine}, to obtain action proposals. Next, the outer-inner-contrastive technique~\cite{shou2018autoloc} is used to calculate the class-specific score for each proposal. To increase the pool of proposals, multiple thresholds are applied, and non-maximum suppression (NMS) is employed to remove duplicate proposals.

\section{Theoretical Derivation} \label{A:Proof}  
Here we provide the derivation of the solution to the following optimal-transport problem in SCC:
\begin{equation} 
	\begin{aligned}\small 
		& \mathop{\min} \ \langle \boldsymbol{Q^S}, -\operatorname{log} \boldsymbol{P^S} \rangle  + \frac{1}{\epsilon} \operatorname{KL} (\boldsymbol{Q^S}, \boldsymbol{\hat{Q}^S}) \quad \textit{s.t.}, \boldsymbol{Q^S} \in \Omega^S \\
	& \Omega^S  = \{ \boldsymbol{Q^S} \in \mathbb{R}_{+}^{N \times K} | \boldsymbol{Q^S} \mathbf{1}^{K}= \boldsymbol{\alpha^S}, {\boldsymbol{Q^S}}^{\top} \mathbf{1}^N=\boldsymbol{\beta^S} \}.
		\label{eq:problem}
	\end{aligned} 
\end{equation}
For notation simplicity, we remove the superscript $S$. Then the problem is rewritten as
\begin{equation} 
	\begin{aligned}\small 
		& \mathop{\min} \ \langle \boldsymbol{Q}, -\operatorname{log} \boldsymbol{P} \rangle  + \frac{1}{\epsilon} \operatorname{KL} (\boldsymbol{Q}||\boldsymbol{\hat{Q}}) \quad \textit{s.t.}, \boldsymbol{Q} \in \Omega \\
	& \Omega  = \{ \boldsymbol{Q} \in \mathbb{R}_{+}^{N \times K} | \boldsymbol{Q} \mathbf{1}^{K}= \boldsymbol{\alpha}, {\boldsymbol{Q}}^{\top} \mathbf{1}^N=\boldsymbol{\beta} \}. 
		\label{eq:problem_simp}
	\end{aligned} 
\end{equation}
To address the problem, we first write the Lagrangian function of~\cref{eq:problem_simp} as follows:
\begin{equation} 
	\begin{aligned}\small 
		\mathcal{L} (\boldsymbol{Q}, \boldsymbol{\mu}, \boldsymbol{\nu})  = & \langle \boldsymbol{Q}, -\operatorname{log} \boldsymbol{P} \rangle  + \frac{1}{\epsilon} \operatorname{KL} (\boldsymbol{Q}||\boldsymbol{\hat{Q}}) \\ & + \boldsymbol{\mu}^\top (\boldsymbol{Q} \mathbf{1}^{K} - \boldsymbol{\alpha}) +\boldsymbol{\nu}^\top ({\boldsymbol{Q}}^{\top} \mathbf{1}^N - \boldsymbol{\beta}) \\
		= & \sum_{n=1}^{N} \sum_{k=1}^{K} (-\boldsymbol{Q}_{n,k} \operatorname{log} \boldsymbol{P}_{n,k} + \frac{1}{\epsilon} \boldsymbol{Q}_{n,k} \operatorname{log} \frac{\boldsymbol{Q}_{n,k}}{\boldsymbol{\hat{Q}}_{n,k} } \\ & + \boldsymbol{\mu}_n \boldsymbol{Q}_{n,k} + \boldsymbol{\nu}_k \boldsymbol{Q}_{n,k} ) - \boldsymbol{\mu}^\top \boldsymbol{\alpha} -\boldsymbol{\nu}^\top \boldsymbol{\beta} 
		\label{eq:problem_lag}
	\end{aligned} 
\end{equation}
where $\boldsymbol{\mu} \in \mathbb{R}^{N} $ and $\boldsymbol{\nu} \in \mathbb{R}^{K} $ are the dual variables so that $\boldsymbol{Q} \mathbf{1}^{K}= \boldsymbol{\alpha}$ and ${\boldsymbol{Q}}^{\top} \mathbf{1}^N=\boldsymbol{\beta}$. The
derivative of $\mathcal{L} (\boldsymbol{Q}, \boldsymbol{\mu}, \boldsymbol{\nu})$ \textit{w.r.t.} $\boldsymbol{Q}_{n,k}$ is:
\begin{equation} 
	\begin{aligned}\small 
	\frac{\partial \mathcal{L} (\boldsymbol{Q}, \boldsymbol{\mu}, \boldsymbol{\nu}) }{\partial \boldsymbol{Q}_{n,k}} = -\operatorname{log} \boldsymbol{P}_{n,k} + \frac{1}{\epsilon} \operatorname{log} \frac{\boldsymbol{Q}_{n,k}}{\boldsymbol{\hat{Q}}_{n,k}} + \frac{1}{\epsilon} + \boldsymbol{\mu}_n  + \boldsymbol{\nu}_k.
	\label{eq:derivative}
	\end{aligned} 
\end{equation}
Note that the optimal $\boldsymbol{Q}$
exists and is unique, as both the objective and the constraint in~\cref{eq:problem_simp} are convex.  Hence, to obtain the optimal $\boldsymbol{Q}$, we set $\frac{\partial \mathcal{L} (\boldsymbol{Q}, \boldsymbol{\mu}, \boldsymbol{\nu}) }{\partial \boldsymbol{Q}_{n,k}} =0$, and then get:
\begin{equation} 
	\begin{aligned}\small 
	\boldsymbol{Q}_{n,k} = e^{- \frac{1}{2} - \epsilon \boldsymbol{\mu}_n - \frac{1}{2} } (\boldsymbol{\hat{Q}}_{n,k}{\boldsymbol{P}^\epsilon_{n,k}}) e^{- \frac{1}{2}- \epsilon \boldsymbol{\nu}_k}.
	\label{eq:derivative_1}
	\end{aligned} 
\end{equation}
Let us denote $\boldsymbol{S} = \boldsymbol{\hat{Q}} \cdot {\boldsymbol{P}^\epsilon}$. Obviously, all elements of $\boldsymbol{S}$ are strictly positive. According to~\cite{sinkhorn1967diagonal,borobia1998matrix,su2017order}, there exist diagonal matrices $\operatorname{diag} (\boldsymbol{u})$ and $\operatorname{diag} (\boldsymbol{v})$ with strictly positive diagonal elements so that $\operatorname{diag} (\boldsymbol{u}) S \operatorname{diag} (\boldsymbol{v})$ belongs to $\Omega$. 

In summary, the optimal $\boldsymbol{Q}$ has the form as: 
\begin{equation} 
	\begin{aligned}\small 
	\boldsymbol{Q} = \operatorname{diag} (\boldsymbol{u}) \boldsymbol{S} \operatorname{diag} (\boldsymbol{v})=\operatorname{diag} (\boldsymbol{u}) (\boldsymbol{\hat{Q}} \cdot {\boldsymbol{P}^\epsilon} )\operatorname{diag} (\boldsymbol{v}),
	\label{eq:solution1}
	\end{aligned}
\end{equation}
where $\boldsymbol{u} \in \mathbb{R}^N$ and $\boldsymbol{v} \in \mathbb{R}^K$ are two renormalization vectors that make the resulting matrix $\boldsymbol{Q}$ to be a probability matrix. 
Throughout our work, we follow~\cite{caron2020unsupervised} to implement the algorithm due to its conciseness. Formally,~\cref{eq:solution1} is replaced as follows:
\begin{equation} 
	\begin{aligned}\small 
	\boldsymbol{Q} =\operatorname{diag} (\boldsymbol{u}) (\boldsymbol{\hat{Q}} \cdot {\exp(\epsilon \boldsymbol{L}}) )\operatorname{diag} (\boldsymbol{v}),
	\label{eq:solution2}
	\end{aligned}
\end{equation}
where $\boldsymbol{L}$ indicates the logits before the $\operatorname{Softmax}$ layer, namely $\boldsymbol{P}=\operatorname{Softmax}(\boldsymbol{L})$. Note that~\cref{eq:solution1} and~\cref{eq:solution2} are equivalent in principle. The main difference lies in the placement of the factor $\epsilon$ that sharpens the labels. In~\cref{eq:solution1}, the factor is applied before $\operatorname{Softmax}$, while in~\cref{eq:solution2}, it is applied after $\operatorname{Softmax}$. Similarly, for the cluster classification component (without $\boldsymbol{\hat{Q}}$), we can obtain the solution as follows:
\begin{equation} 
	\begin{aligned}\small 
	\boldsymbol{Q} =\operatorname{diag} (\boldsymbol{u}) {\exp(\epsilon \boldsymbol{L}}) \operatorname{diag} (\boldsymbol{v}).
	\label{eq:solution3}
	\end{aligned}
\end{equation}
Both~\cref{eq:solution2} and~\cref{eq:solution3} can be efficiently computed using the iterative Sinkhorn-Knopp algorithm~\cite{cuturi2013sinkhorn}. We refer to~\cite{cuturi2013sinkhorn} for more details. This algorithm is highly efficient on GPU as it only involves a couple of matrix multiplication, enabling online computation.

\end{document}